\documentclass{article}

\usepackage[nonatbib,final]{neurips_2020}

\usepackage{amsmath,amsthm,wrapfig}
	\usepackage[utf8]{inputenc} 
\usepackage{amsmath, amssymb,bm, cases, mathtools, thmtools}
\usepackage{verbatim}
\usepackage{graphicx}\graphicspath{{figures/}}
\usepackage{multicol}
\usepackage{tabularx}
\usepackage[usenames,dvipsnames]{xcolor}
\usepackage{mathrsfs} 
\usepackage{caption}
\usepackage{algorithm}
\usepackage{algorithmicx}
\usepackage[noend]{algpseudocode}

\usepackage[%
    minnames=1,maxnames=99,maxcitenames=3,
    style=numeric,%
    sorting=none,
    doi=false,url=false,
    giveninits=true,
    hyperref,natbib,backend=biber]{biblatex}
\renewbibmacro{in:}{%
  \ifentrytype{article}{}{\printtext{\bibstring{in}\intitlepunct}}}
\bibliography{biblio}

\usepackage[T1]{fontenc}
\usepackage{inconsolata}
\usepackage{mathptmx}

\usepackage{listings} %

\usepackage{enumitem}
\usepackage{booktabs}       %
\usepackage{nicefrac}       %

\usepackage{lineno}
\usepackage{bbm}

\usepackage{caption}
\usepackage{subcaption}

\usepackage{datetime}

\DeclareMathAlphabet\EuRoman{U}{eur}{m}{n}
\SetMathAlphabet\EuRoman{bold}{U}{eur}{b}{n}

\makeatletter
\let\reftagform@=\tagform@
\def\tagform@#1{\maketag@@@{\ignorespaces\textcolor{gray}{(#1)}\unskip\@@italiccorr}}
\renewcommand{\eqref}[1]{\textup{\reftagform@{\ref{#1}}}}
\makeatother

\usepackage{xparse}
\usepackage{xstring}

\def\[#1\]{\begin{align}#1\end{align}}
\def\*[#1\]{\begin{align*}#1\end{align*}}

\newcommand{\testset}{S^{\textrm{test}}}

\newcommand{\minibatch}{\bar{S}}

\newcommand{\ip}[2]{\langle #1,#2\rangle}

\newcommand\optparen[1]{\ifthenelse{\equal{#1}{}}{}{(#1)}}
\newcommand{\RiskChar}{R}
\newcommand{\Risk}[2]{\RiskChar_{#1}\optparen{#2}}
\newcommand{\Error}[2]{\RiskChar^{0-1}_{#1}\optparen{#2}}
\newcommand{\EmpRisk}[2]{\hat \RiskChar_{#1}\optparen{#2}}

\newcommand{\Reals}{\mathbb{R}}

\newcommand{\Nats}{\mathbb{N}}

\newcommand{\grad}{\nabla}

\newcommand{\dee}{\mathrm{d}}

\DeclareMathOperator*{\newlim}{\mathrm{lim}\vphantom{\mathrm{infsup}}}

\DeclareMathOperator*{\newmax}{\mathrm{max}\vphantom{\mathrm{infsup}}}

\renewcommand{\lim}{\newlim}

\renewcommand{\max}{\newmax}

\newcommand{\norm}[1]{\lVert #1 \rVert}

\newcommand{\tuple}[1]{\langle #1 \rangle}

\newcommand{\trace}{\mathrm{Tr}}

\newcommand{\loss}{\ell}

\usepackage[hide=true,setmargin=true,marginparwidth=1.25in]{marginalia}

\usepackage[utf8]{inputenc} %
\usepackage[T1]{fontenc}    %
\usepackage{url}            %
\usepackage{booktabs}       %
\usepackage{amsfonts}       %
\usepackage{nicefrac}       %
\usepackage{microtype}      %

\usepackage{amsmath}
\usepackage{graphicx,adjustbox,crop}

\usepackage{xcolor,xspace}

\usepackage[colorlinks,citecolor=blue,urlcolor=blue,linkcolor=RawSienna]{hyperref}

\usepackage[capitalize]{cleveref}

\crefname{lemma}{Lemma}{Lemmas}
\crefname{corollary}{Corollary}{Corollaries}
\crefname{theorem}{Theorem}{Theorems}

\newcommand{\Taylorized}{Linearized\xspace}

\title{Deep learning versus kernel learning: an empirical study of loss landscape geometry and the time evolution of the Neural Tangent Kernel}

\newcommand*\samethanks[1][\value{footnote}]{\footnotemark[#1]}

\author{Stanislav Fort$^1$\thanks{Equal contributions. Correspondence to: \texttt{sfort1@stanford.edu}, \texttt{karolina.dziugaite@elementai.com} }\ \ \ \ \ \ \     Gintare Karolina Dziugaite$^2$\samethanks{}\  \ \ \ \ \ \   Mansheej Paul$^1$ \\ \textbf{Sepideh Kharaghani$^2$\ \ \ \ \ \ \  Daniel M. Roy$^{3,4}$\ \ \ \ \ \ \  Surya Ganguli$^1$}\\
$^1$Stanford University\ \ \ \  $^2$Element AI\ \ \ \  $^3$University of Toronto\ \ \ \  $^4$Vector Institute  }

\begin{document}

\maketitle

\begin{abstract}

In suitably initialized wide networks, small learning rates transform deep neural networks (DNNs) into neural tangent kernel (NTK) machines, whose training dynamics is well-approximated by a linear weight expansion of the network at initialization. 
Standard training, however, diverges from its linearization in ways that are poorly understood. 
We study the relationship between the training dynamics of nonlinear deep networks, the geometry of the loss landscape, and the time evolution of a data-dependent NTK. 
We do so through a large-scale phenomenological analysis of training, synthesizing diverse measures characterizing loss landscape geometry and NTK dynamics. 
In multiple neural architectures and datasets, we find these diverse measures evolve in a highly correlated manner, revealing a universal picture of the deep learning process. 
In this picture, deep network training exhibits a highly chaotic rapid initial transient that within 2 to 3 epochs determines the final linearly connected basin of low loss containing the end point of training.  
During this chaotic transient, the NTK changes rapidly, learning useful features from the training data that enables it to outperform the standard initial NTK by a factor of 3 in less than 3 to 4 epochs.  
After this rapid chaotic transient, the NTK changes at constant velocity, and its performance matches that of full network training in 15\% to 45\% of training time.   
Overall, our analysis reveals a striking correlation between a diverse set of metrics over training time, governed by a rapid chaotic to stable transition in the first few epochs, that together poses challenges and opportunities for the development of more accurate theories of deep learning.

\end{abstract}

The remarkable empirical success of deep learning across a range of domains stands in stark contrast to our theoretical understanding of the mechanisms underlying this same success \cite{Bahri2020-mi}. Indeed, we are currently far from a mature, unified mathematical theory of deep learning that is powerful enough to universally guide engineering design choices.  
As in many other fields of inquiry, a key prerequisite to any such theory is careful empirical measurements of the deep learning process, with the scientific aim of unearthing combinations of variables that obey correlated dynamical laws that can serve as the inspiration for future theories.  
Indeed, a large body of work has studied, mainly in isolation, diverse and intriguing phenomenological properties, as well as extreme simplifying theoretical limits, of deep learning. 
In particular, we focus on $3$ intertwined aspects of deep learning that have previously been studied largely in isolation: (1) the large scale structure of deep learning loss surfaces, (2) the local geometry of such loss surfaces, and (3) and the performance of linearized training methods, like the neural tangent kernel (NTK), that has gained attention through its ability to theoretically describe an infinite width low learning rate limit of deep learning in terms of kernel machines with random data-independent kernels.  
The fundamental goal of this work is to obtain a more integrative view of the intertwined relations between loss landscape geometry at multiple scales of organization and the dynamics of learning in deep networks, by performing {\it simultaneous} measurements of many diverse properties.  
We describe the previous work that motivates our current measurements in \cref{sec:diverseaspects}, and we summarize our results and contributions in \cref{sec:discussion}, which can be read right after \cref{sec:diverseaspects}.

\vspace{-0.3cm}
\section{Diverse aspects of deep learning phenomenology}
\vspace{-0.3cm}
\label{sec:diverseaspects} 

\begin{figure}[t!]
\centering
\includegraphics[
                 width=1\linewidth
                ]{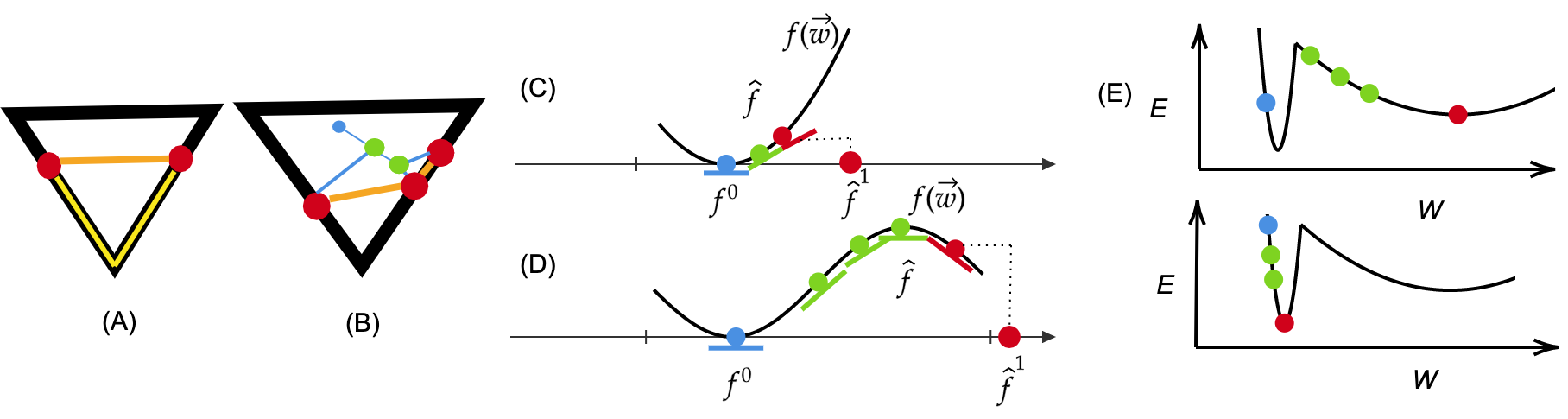}
\caption{A conceptual overview of diverse deep learning phenomenology. 
(A) A schematic picture of the region of low loss (black area) in weight space as a network of high dimensional basins with lower dimensional intersections, motivated by recent work \citep{li2018measuring,goldilocks,draxler2018essentially,garipov2018loss,fort2019large}. 
Two networks (red points) in different basins can be connected by a low loss nonlinear path (yellow) but not by a low loss linear path (orange). 
(B) A schematic view of the process of hierarchically exploring loss landscapes by spawning child networks \citep{frankle2019linear}. 
A randomly initialized parent network (blue point) is trained up to a certain spawn epoch (green point) at which two (or more) child networks are spawned from with identical weights and then subsequently trained independently with different SGD minibatches (bifurcating blue lines).  
Two children spawned later (earlier) than a very early transition time in parent training, will arrive at the same (different) basin on the loss landscape. 
(C) A schematic view of NTK training. 
The black curve is the space of functions $f_w$ realizable by varying the parameters $w$ of a neural network and full network training proceeds along this curved function manifold (blue to green to red points). NTK training linearizes the manifold at initialization (blue point), and trains along the tangent space (blue line). Such linearized training is equivalent to kernel regression in function space where the kernel is closely related to the tangent plane along which training occurs. 
This panel shows a case where NTK and full nonlinear training are similar in that the kernel at initialization does not change much over learning, as shown schematically by the similar orientations of the initial (blue), intermediate (green) and final (red) tangent spaces.  
(D) The same as in panel (C), except now showing schematically a case where the NTK method is very different from full nonlinear training, in which the kernel changes considerably, as evidenced by the strong twisting of tangent spaces (blue, green and red lines), resulting in a final learned kernel (associated with the red tangent space) that is quite different from the initial random kernel (associated with the blue tangent space). 
(E) Consider an error landscape with a sharp and a wide minimum separated by an error barrier.  With a small learning rate (bottom), a learning trajectory starting at an initial point (blue) will slowly descend through intermediate points (green) to a minimum position (red) in the sharp minimum, and is unable to escape it. 
With a larger learning rate (top), a learning trajectory that starts in the sharp minimum 
at a position (blue point) that is even \emph{lower} than the error barrier, can escape the 
sharp minimum.}
\vspace{-0.3cm}
\label{fig:conceptual}
\end{figure}

\paragraph{The large scale geometric structure of neural loss landscapes.} Recent work 
has revealed many insights into the shape of loss functions over the high dimensional space of neural network parameters.  
For example, \citep{li2018measuring,goldilocks} demonstrates 
that training even within a random, low-dimensional affine subspace of parameter space can yield a network with low test loss. This suggests that the region of parameter space with low test loss must be a relatively high dimensional object, such that low dimensional random affine hyperplanes can generically intersect it.  
Moreover, \citep{draxler2018essentially,kuditipudi2019explaining,garipov2018loss} show that different, independently trained networks in weight space with low loss can be connected through nonlinear pathways (found via an optimization process) that never leave this low loss manifold.  
However,
\emph{direct linear} pathways connecting two such independently trained networks typically always leave the low loss manifold.  
The loss function restricted to such linear paths then yields a loss barrier at an intermediate point between the two networks. 
\citep{fort2019large} builds and provides evidence for a unifying geometric model of the low-loss manifold consisting of a network of mutually intersecting high dimensional basins (\cref{fig:conceptual}A). Two networks within a basin
can be connected by a straight line that never leaves the low-loss manifold, while two networks in different basins can be connected by a piecewise linear path of low loss that is forced to traverse the intersection between two basins.  
\cite{fort2019deep} uses these insights to argue that deep ensembles are hard to beat using local subspace sampling methods due to the geometry of this underlying loss landscape. 
\cite{frankle2019linear} provides further evidence for this large-scale structure by demonstrating that {\it after} a very early stage of training of a parent network (but not earlier) two child networks trained starting from the parameters of the parent end up in the same low loss basin at the end of training, and could be connected by a {\it linear} path in weight space that does not leave the low loss manifold (\cref{fig:conceptual}B).  
Furthermore, \cite{jastrzebski2020breakeven,leclerc2020two} show that the properties of the final minimum found are strongly influenced by the very early stages of training.  
Taken together, these results present an intriguing glimpse into the large scale structure of the low loss manifold, and the importance of early training dynamics in determining the final position on the manifold.

\vspace{-0.3cm}
\paragraph{Neural tangent kernels, linearized training and the infinite width limit.} The neural tangent kernel (NTK) has garnered much attention as it provides a theoretical foothold to understand deep networks, at least in an infinite width limit with appropriate initialization scale and low learning rate \cite{jacot2018neural,novak2019neural}.  
In such a limit, a network does not move very far in weight space over the course of training, and so one can view learning as a linear process occurring along the tangent space to the manifold of functions $f_w$ realizable by the parameters $w$, at the initial function $f^0$ (\cref{fig:conceptual}C). 
This learning process is well described by kernel regression with a certain random kernel associated with the tangent space at initialization. 
The NTK is also a special case of Taylorized training \cite{bai2020taylorized}, which approximates the realizable function space $f_w$ to higher order in the vicinity of initialization. 
Various works compare the training of deep networks to the NTK \cite{arora2019exact,lee2019wide,arora2019harnessing,shankar2020neural}. 
In many cases, state of the art networks outperform their random kernel counterparts by significant margins, suggesting that deep learning in practice may indeed explore regions of function space far from initialization, with the tangent space twisting significantly over training time, and hence the kernel being learned from the data (\cref{fig:conceptual}D). However, the nature and extent of this function space motion, the degree of tangent space twisting, and how and when data is infused into a {\it learned} tangent kernel, remains poorly understood.  
 
\vspace{-0.3cm}
\paragraph{The local geometric structure of neural loss landscapes.}  
Much effort has gone into characterizing the local geometry of loss landscapes in terms of Hessian curvature and its impact on generalization and learning.  
Interestingly \cite{papyan2019measurements} analyses the Hessian eigenspectrum of loss landscapes at scale, demonstrating that learning leads to the emergence of a small number of large Hessian eigenvalues, and many small ones, bolstering evidence for the existence of many flat directions in low loss regions depicted schematically in \cref{fig:conceptual}A.
\cite{fort2019emergent} shows that the gradients of logits with respect to parameters cluster tightly based on the logit over training time, leading directly to the emergence of very sharp Hessian eigenvalues.  
Moreover, a variety of work has explored relations between the curvature of local minima found by training and their generalization properties \citep{dziugaite2017computing,chaudhari2019entropy,langford2002bounding,hochreiter1997flat,keskar2016large,fort2019large, fort2018goldilocks}, and how learning rate and batch size affect the curvature of the minima found \citep{jastrzbski2017factors, sagun2017empirical, NIPS2018_8049},  with larger learning rates generically enabling escape from sharper minima (\cref{fig:conceptual}E).  
\cite{lewkowycz2020large} makes a connection between learning rates and the validity of NTK training, showing that for infinitely wide networks, training with a learning rate above a scale determined by the top eigenvalue of the Hessian at initialization results in a learning trajectory that outperforms NTK training, presumably by exploring nonlocal regions of function space far away from initialization. 

\vspace{-0.3cm}
\paragraph{Towards an integrative view.} Above, we have reviewed previously distinct strands of inquiry into deep learning phenomenology that have made little to no contact with each other.  
Indeed, we currently have no understanding of how local and global loss geometry interacts with the degree of kernel learning in state of the art architectures and training regimes used in practice.  
For example, at what point in training is the fate of the final chosen basin in \cref{fig:conceptual} A,B irrevocably determined?  
Does the kernel change significantly from initialization as in \cref{fig:conceptual}D?
If so, when during training does the tangent kernel start to acquire knowledge of the data?
Also, when does kernel learning finally stabilize? 
What relations do either of these times have to the time at which basin fate is determined?
How does local geometry in terms of curvature change as all these events occur? Here we address these questions to obtain an integrative view of the learning process across a range of networks and datasets. 
While we only present results for ResNet20 trained on CIFAR10 and CIFAR100 in the main paper, in \cref{app:additionalfigs} we find similar results for a WideResNet, variations of Resnets and a Simple CNN trained on CIFAR10 and CIFAR100, 
indicating our results hold generally across architectures, datasets and training protocols. Many experimental details are covered in our Appendix.

\vspace{-0.3cm}
\section{Definition of measurement metrics for geometry and training}
\vspace{-0.3cm}
\label{sec:preliminaries}

We now mathematically formalize the quantities introduced in the previous section as well as define more quantities whose dynamics we will measure during training.  Let $S = ((x_i,y_i), 1 \le i \le m)$ be $m$ training examples, with $y_i \in \{0,1\}^K$, where $K$ is the number of classes.
Let $f_{w}(x)$ denote the $K$-dimensional output vector of logits, of a neural network parameterized by weights $w \in \Reals^d$ on input $x$. 
We are interested in the average classification error $\Error{S}{w}$ over the samples $S$. 
For training purposes, we also consider a (surrogate) loss $\loss(\hat y,y)$ for predicting $\hat y$ when the true label is $y$. 
Denote by $g(\hat y,y)$  the gradient of $y' \mapsto \loss(y',y)$, evaluated at $y'=\hat y$. 
Write $g_{w}(S)$ for concatenation of the gradient vectors $g(f_{w}(x_i),y_i)$, for $i=1,\dots,m$. 
Let $J_w(x) \in \Reals^{K \times d}$ be the Jacobian of $f_w(x)$ with respect to the parameters $w$. 
Define $J_w (S) \in \Reals^{ m K \times d}$ to be the concatenation of $J_w(x_1),\dots,J_w(x_m)$, which is then the Jacobian of $f_w(S)$ with respect to the parameters $w$. 
The $k^{th}$ row of $J_w(x)$,  denoted $(J_w(x))_k$, is a vector in $\Reals^d$.
Let $H_w(x)$ be the $K \times d \times d$ tensor where $(H_w(x))_k = \grad_w (J_w (x))_{k} \in \Reals^{d \times d}$ is the Hessian of logit $k$ w.r.t. weights $w$. 

\vspace{-0.3cm}
\paragraph{Training Dynamics, \Taylorized training, and introduction of a data-dependent NTK.} Let $(w_t)_{t \in \Nats}$ be the weights at each iteration of SGD, based on minibatch estimates of the training loss $ \EmpRisk{\minibatch}{w} = \frac{1}{n} \sum_{x_i \in \minibatch} \loss(f_{w}(x_i),y_i)$, 
where $\minibatch \subset S$ is a subsample of data of size $n$. We write $f_t(x)$ for $f_{w_t}(x)$
and similarly for $g_t$, $J_t$, and $H_t$.
The SGD update with learning rate $\eta$ is then 
\begin{align}\label{sgdupdate}
\Delta_t 
:= w_{t+1} - w_t &= - \eta \grad_{w} \Risk{\minibatch}{w_t},
\end{align}
Consider also a second-order Taylor expansion to approximate the change to the logits for input $x_i$:
\[\label{eq:taylorsecond}
f_{t+1}(x_i) 
 \approx f_{t}(x_i) 
       -  J_{t}(x_i) \Delta_t + 
       \norm{\Delta_t}_{H_t(x_i)},
\]
where
\[\label{secondordercond}
\norm{ \Delta_t }_{(H_t(x_i))_k^T } := \ip{\Delta_t}{(H_t(x_i))_k^T \Delta_t}.
\]

Note, that for an infinitesimal $\eta$, the dynamics in \cref{sgdupdate} are those of gradient flow, and terms higher than order 1 in \cref{eq:taylorsecond} vanish. 
In this case, steepest descent in the parameter space corresponds to steepest descent in the function space using a \emph{neural tangent kernel} (NTK), 
\[
\kappa_t(x,x') = J_t(x) J_t(x')^T.
\]
Let $\kappa_t(S)$ denote the $m$ by $m$ gram matrix with $i,j$ entry
$\kappa_t(x_i,x_j)$.
If $\kappa_t(S) = \kappa_{t_0}(S)$ for $t>t_0$, i.e., if the tangent kernel is constant over time, then the dynamics correspond to those of training the neural network linearized at time $t_0$. 
The kernel has been shown to be nearly constant in the case of very wide neural networks at initialization (see, e.g., \citep{jacot2018neural,zou2019improved,ji2019polylogarithmic,du2018gradient,lee2019wide,chen2019much}).
Intuitively, we can think of each of the $d$ columns of $J_w(x) \in \Reals^{K \times d}$ as a tangent vector to the manifold of realizable neural network functions in the ambient space of all functions of $K$ logits over input space $x$, at the point $f_w(x)$ in function space. 
Thus the span of the $d$ columns of $J_w(x)$, as $x$ varies, constitute the tangent planes in function space depicted schematically in \cref{fig:conceptual}CD.
Since the kernel is the Gram matrix associated with these tangent functions, evaluated at the training points, then if the tangent space twists substantially, the kernel necessarily changes (as in \cref{fig:conceptual}D). 

Conversely, if the NTK does not change substantially from initialization, then the full SGD training can be well approximated by training along the tangent space to $f_0$ at initialization, yielding the linearized training dynamics.  
This approach can be generalized to training along higher order Taylor approximations of the manifold $f_w(x)$ in the vicinity of the initial function $f_0$ \cite{bai2020taylorized}.  
In this work, in order to explore function space geometry and its impact on training, we extend this approach by doing full network training up to time $\tilde t$, and then linearized training subsequently. 
This yields a linearized training trajectory $\{w^{\tilde{t}}_t \}_{t= \tilde{t}}^{T}$, which can then be compared to the weight dynamics under full training (see Appendix for details). 
This approach geometrically corresponds to training along an intermediate tangent plane (one of the green planes in \cref{fig:conceptual}CD), or equivalently, corresponds to learning with a \emph{data-dependent} NTK.  
This novel examination of how much training time is required to learn a high performing NTK, distinct from the random one used at initialization, and relations between this time and both the local and large scale structure of the loss landscape, constitutes a key contribution of our work.

\vspace{-0.3cm}
\paragraph{Hierarchical exploration of the loss landscape through parents and children.}
In order to explore the loss landscape and the stability of training dynamics in a more multiscale hierarchical manner than is possible using completely independent training runs, we employ a method of parent-child spawning \cite{frankle2019linear} (shown schematically in \cref{fig:conceptual}B). 
In this process, a parent network is trained from initialization to a spawning time $t_s$, yielding a parent weight trajectory $\{w_t\}_{t=0}^{t_s}$. 
At the spawn time $t_s$, several copies of the parent network are made, and these so-called children are then trained with independent minibatch stochasticity, yielding different child weight trajectories $\{w^{t_s,a}_t\}_{t=t_s}^{T}$, where $a$ indexes the children, and $T$ is the final training time. 
We will be interested in various measures of the distance between children after training, as a function of their spawn time $t_s$, as well as measures of the distance between the same network (either parent or child) at two different training times.  We turn to these various distance measures next.

\vspace{-0.3cm}
\paragraph{Kernel distance.}

For finite width networks, the kernel $\kappa_{t}(S)=\kappa_{w_t}(S)$ changes with training time $t$. We compare two Kernel gram matrices in a scale-invariant manner by computing a \emph{kernel distance}:
\[\nonumber\textstyle
S(w,w') = 1-\frac{ \trace  (\kappa_w(S) \kappa_{w'}^T(S) ) }{ \sqrt{ \trace (\kappa_w(S) \kappa_{w}^T(S) ) } \sqrt{ \trace (\kappa_{w'}(S) \kappa_{w'}^T(S) ) }}.
\]

\vspace{-0.3cm}
\paragraph{Kernel velocity.}

We further track the speed at which the kernel changes. As discussed above, in non-linear neural networks, we do not expect  \cref{secondordercond} to vanish. In order to capture the evolution of the quantity in \cref{secondordercond}, we compute the \emph{kernel velocity} $v(t) \equiv S(w_t,w_{t+dt})  / dt$, i.e. the rate of change of kernel distance. We use a time separation of $0.4$ epochs to capture appreciable change.

\vspace{-0.3cm}
\paragraph{Error barrier between children.}
To assess (and indeed \emph{define}) whether two children arrive at the same basin or not at the end of training (see e.g. \cref{fig:conceptual}AB), we compute the error barrier between children along a \emph{linear} path interpolating between them in weight space. 
Let $w^{\alpha} _t =  \alpha w_t + (1-\alpha) w'_t$, where $w'_t$ and $w_t$ are the weights of two child networks, spawned from some iteration $t_s$, and $\alpha \in [0,1]$.
At various $t_s$ we compute $\max_{ \alpha \in [0,1] } \EmpRisk{S}{w^{\alpha} _t } - \frac1 2 {\big(\EmpRisk{S}{w_t} + \EmpRisk{S}{w'_t}\big)}$, 
which we call the \emph{error barrier}. 
Note, that the error barrier at the end of training between two children is the same as \emph{instability} in \citep{frankle2019linear}.

\vspace{-0.3cm}
\paragraph{ReLU activation pattern distance.}
In a ReLU network, the post-nonlinearity activations in layer $l$ are either greater or equal to $0$.
We can thus construct a tensor $B_w(S)$, with $(B_w(S))^{i,j,l} = 1$ if for an input $x_i$, $j^\mathrm{th}$ node in the $l^\mathrm{th}$ layer is strictly positive, and $(B_w(S))^{i,j,l} = 0$ otherwise. We compare ReLU on/off similarity between networks parameterized by $w$ and $w'$ by computing the Hamming distance between $B_w(S)$ and $B_{w'}(S)$, and normalizing by the total number of entries in $B_w(S)$.

\begin{figure}[h]
    \centering
    \includegraphics[width=\linewidth]{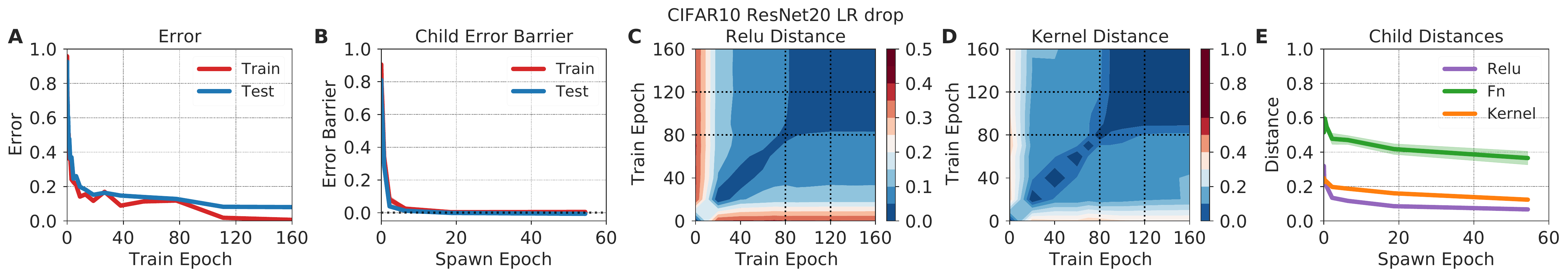}
    \caption{SOTA ResNet20 trained on CIFAR10 using SGD with momentum and learning rate drops. 
    }
    \label{fig:megaplot1}
    \centering
    \includegraphics[width=\linewidth]{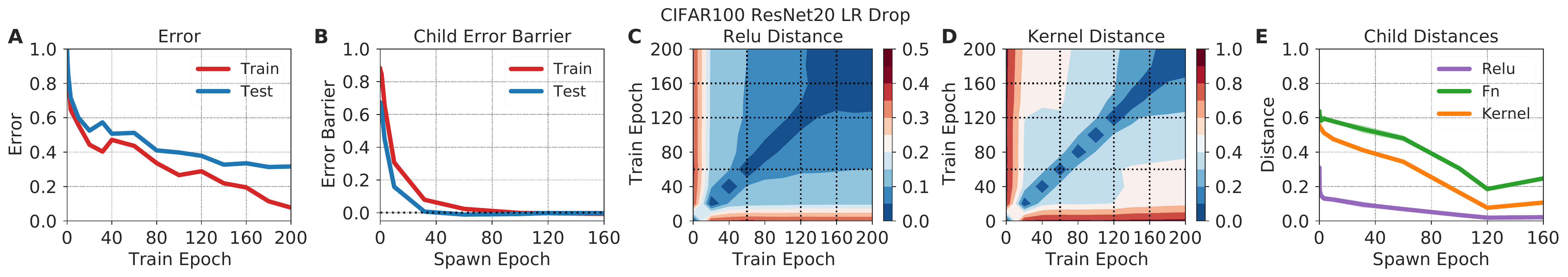}
    \caption{ResNet20 trained on CIFAR100 using SGD with momentum and learning rate drops. 
    }
    \label{fig:megaplot2}
    \centering
    \includegraphics[width=\linewidth]{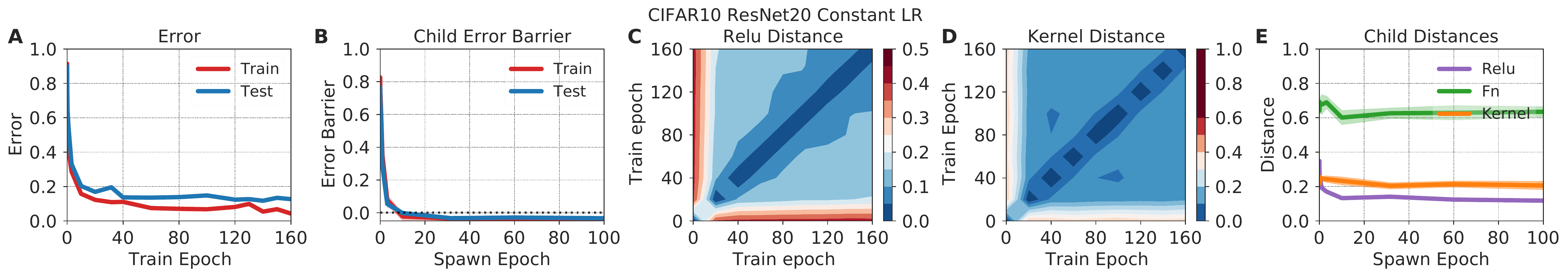}
    \caption{ResNet20 trained on CIFAR10 using SGD with momentum and constant learning rate.
    \ \\ \ \\
    \Cref{fig:megaplot1,fig:megaplot2,fig:megaplot0}: An integrated view of learning. 
    (A) Parent network learning curves. 
    (B) Error barrier between pairs of children at the end of training, as a function of spawn time, with children trained for same number of epochs as the parent. 
    (C) and (D) Heatmaps representing the ReLU and kernel distance between a parent network at different pairs of training times. Dashed black lines indicate epochs at which the learning rate is dropped. (E) ReLU, function space, and kernel distances between pairs children at the end of training, as a function of spawn time. }
    \label{fig:megaplot0}
\end{figure}

\vspace{-0.3cm}
\paragraph{Function space distance.}
To compute the distance between the two functions $f_w$ and $f_{w'}$, parameterized by weights $w$ and $w'$, we would ideally like to calculate the degree of disagreement between their outputs averaged over the whole input space $x$.  
However, since this is computationally intractable, we approximate this distance by the normalized fraction of test examples on which their predicted labels disagree. 
Let $\testset$ denote the test set. 
Then,
$
    \norm{f_w(x) - f_{w'}(x)}_{\testset} = 
    \frac{1}{Z |\testset_x|}\sum_{x \in \testset_x} \left(f_w(x) \neq f_{w'}(x) \right)
$ l
where $\testset_x$ are test inputs and $Z$ is a normalizing constant chosen to aid comparison. In particular, we define $Z$ to be 
the expected number of examples on which two classifiers 
would disagree assuming each made random independent predictions with the same error rates, $p$ and $p'$, as their error rates on the test set. This quantity is used also by \citep{fort2019deep},
and is given by $Z = p(1-p') + p'(1-p) + pp'\frac{K-2}{K-1}$, where $K$ is the number of classes.
 A unit distance indicates two networks make uncorrelated errors.

\vspace{-0.3cm}
\section{An integrative view of learning dynamics}
\vspace{-0.3cm}

\cref{fig:megaplot1,fig:megaplot2} plot the full range of metrics defined in \Cref{sec:preliminaries} for two SOTA networks.
Panel A presents standard training curves.
Panel B confirms the results of \citep{frankle2019linear}, that the error barrier on a linear path between two children decreases rapidly with spawning time, falling close to $0$ within two to five epochs. 
Panel C and D indicate that the NTK changes rapidly early in training, and more slowly later in training,  as quantified by ReLU activation distance (C) and kernel distance (D) measured on a parent run at different pairs of times in training.
Finally, Panel E shows that function, kernel and ReLU distances between children at the end of training also drop as a function of spawn time.

We note that the SOTA training protocols in \cref{fig:megaplot1,fig:megaplot2} involve learning rate drops later in training, which alone could account for a slowing of the NTK evolution. Therefore we ran a constant learning rate experiment in \cref{fig:megaplot0}. We see that all tracked metrics still exhibit the same key patterns: the error barrier drops rapidly within a few epochs (B), the NTK evolves very rapidly early on, but continues to evolve at a constant slow velocity later in training (C,D), and final distances between children drop at an early spawn time and remain constant thereafter (E).  

Overall, these results provide an integrative picture of the learning process, which reveals an early, extremely short chaotic period in which the final basin chosen by a child is highly sensitive to SGD noise and the NTK evolves very rapidly, followed by a later more stable phase in which the basin selection is determined, the NTK continues to evolve, albeit more slowly, and the final distance between children remains smaller.  In the next few sections we explore these results in more detail.

Equivalent results for other networks are shown in \cref{app:additionalfigs} in \cref{fig:megaplot3,fig:megaplot4,fig:megaplot5,fig:megaplot6,fig:megaplot7,fig:megaplot8}, together with additional network properties tracked over epochs.

\vspace{-0.3cm}
\section{The local and global geometry of the loss landscape surrounding children}
\label{sec:childgeo} 
\begin{figure}[h]
\centering
\includegraphics[width=0.9\linewidth]{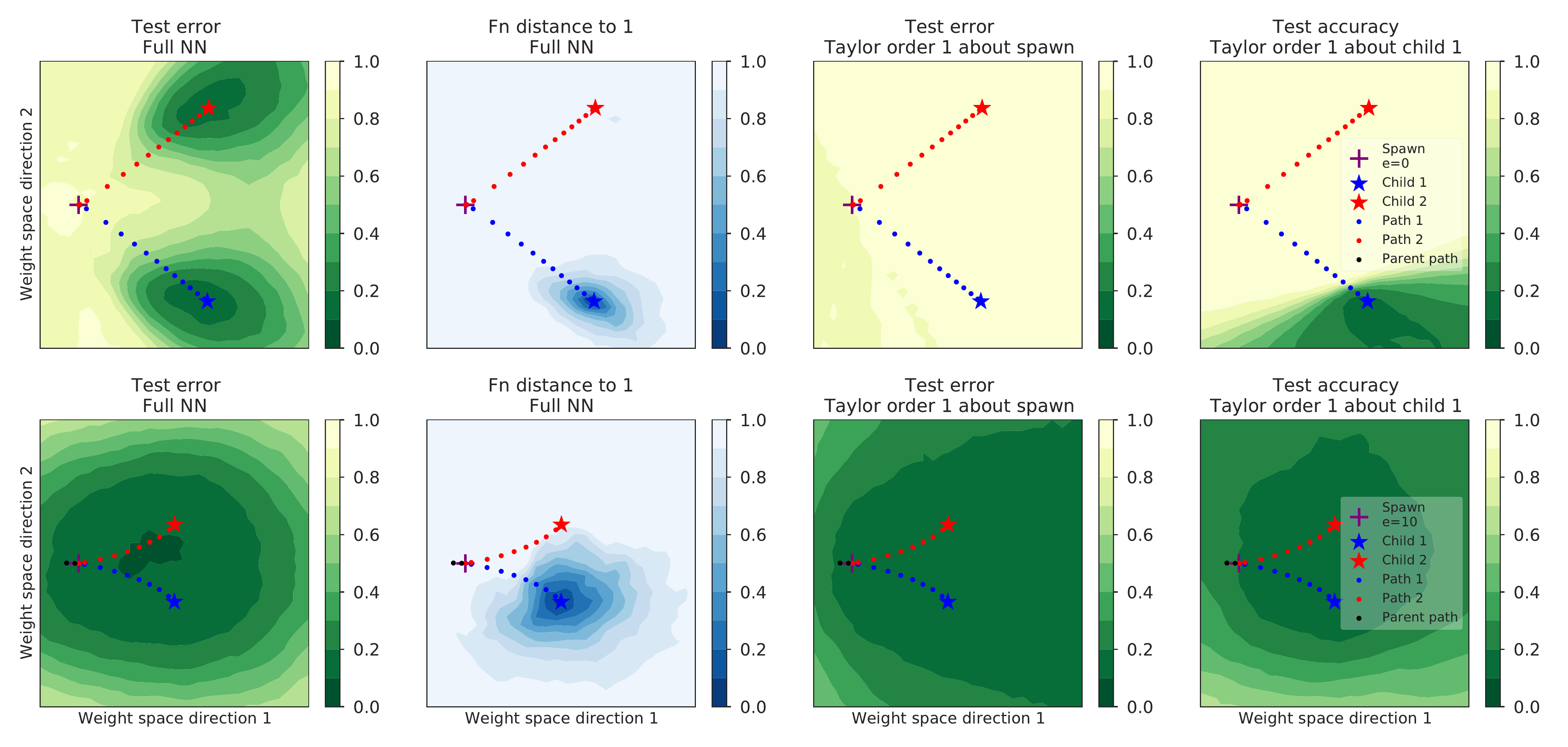}
\caption{The error landscape and function space geometry on a 2D section defined by a pair of children (red and blue stars) and the spawning parent (purple cross) when the spawn point is in the early chaotic (top row) and late stable (bottom row) regimes of training.
All other training points are projected to this $2D$ section. 
The left two columns show, as a function of position on this $2D$ section, the test error and the function space distance to a chosen child (blue star). 
The right two columns show the test error along an affine tangent plane in function space obtained by a first order Taylor expansion of $f_w$
in weight space around the weights of two different networks (the spawning parent and one of the children).  
A function space point along the tangent plane at $f_w$ is identified with a point on the curved 2D section in function space through the relation $f_w + \Delta w \cdot \grad_w f_w  \rightarrow  f_{w+\Delta w}$. 
}
\label{fig:children_spreading}
\vspace{-0.3cm}
\end{figure}

\vspace{-0.3cm}
We first explore how both the global and local landscape geometry surrounding two child pairs and their spawning parent depend on the spawn time $t_s$ in \cref{fig:children_spreading}. These three networks define a $2D$ affine plane in weight space and a curved $2D$ manifold in function space.
The first two columns of \cref{fig:children_spreading} clearly indicate that two children spawned at an early time $t_s$ in the chaotic training regime arrive at two different loss basins that are well separated in function space (top row), while two children spawned at a later time $t_s$ in the stable training regime arrive at the same loss basin, though this loss basin can still exhibit non-negligible function diversity (albeit smaller than the diversity between basins).  
Furthermore, the right two columns of \cref{fig:children_spreading} indicate that the test error as a function of position along the tangent plane to the 2D curved manifold in function space (either at the spawn point or a child point) is insufficient to describe the error along the full curved $2D$ manifold in function space when the children are in different basins (top row), but can approximately describe the loss landscape when the children are in the same basin (bottom row).  
Thus \cref{fig:children_spreading} constitutes a new direct \emph{data-driven} visualization of loss landscape geometry that provides strong evidence for several aspects of the conceptual picture laid out in \cref{fig:conceptual}: the existence of multiple basins (\cref{fig:conceptual}A), the chaotic sensitivity of basin fate to SGD choices early in training (\cref{fig:conceptual}B), and the twisting of tangent planes in function space that occur as one travels from one basin to another (\cref{fig:conceptual}ACD). See also \cref{fig:tsne_bn_main} for a t-SNE visualization of the bifurcating evolution of all parents and children in function space that further corroborates this picture of loss landscape geometry and its impact on training dynamics.

\begin{figure}[h]
\centering
\includegraphics[width=0.46\linewidth]{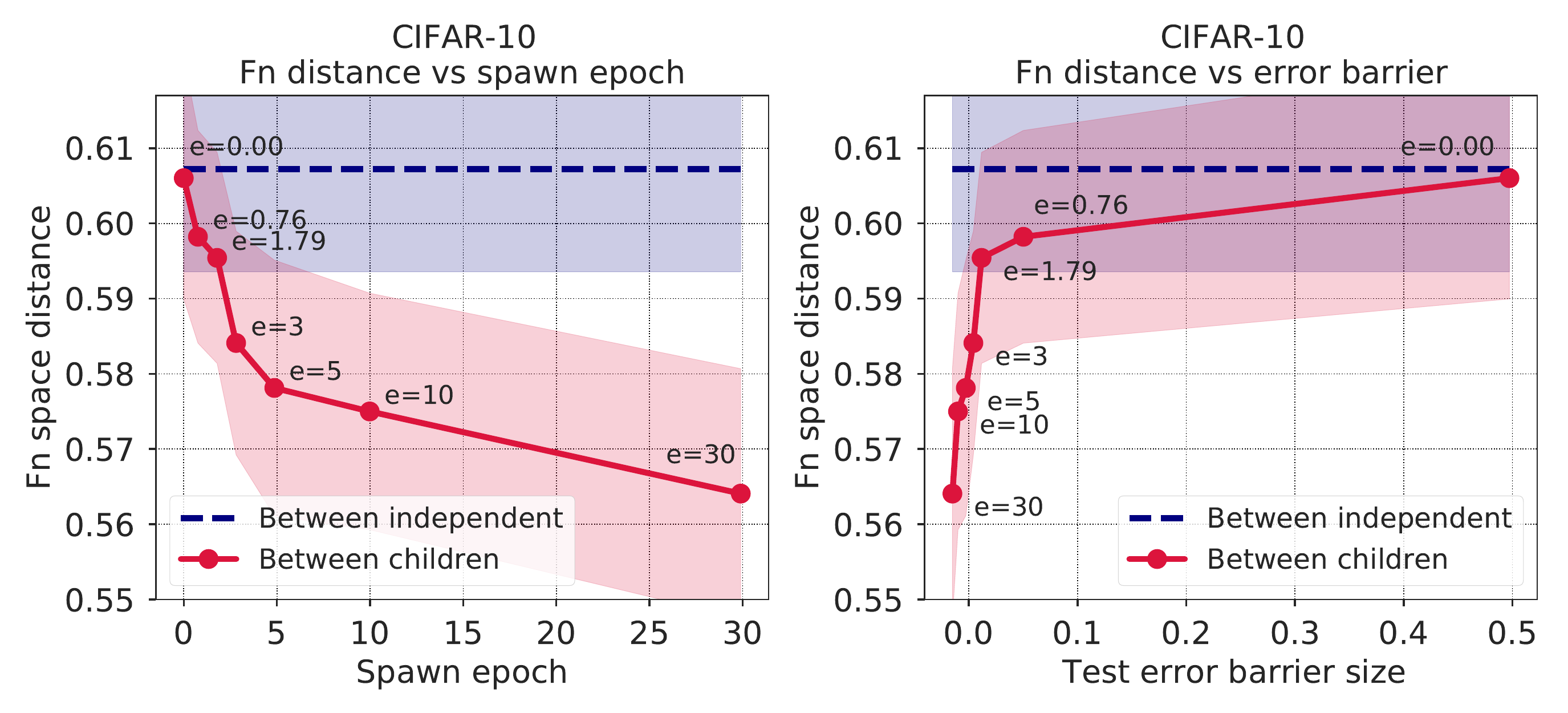}
\includegraphics[width=0.46\linewidth]{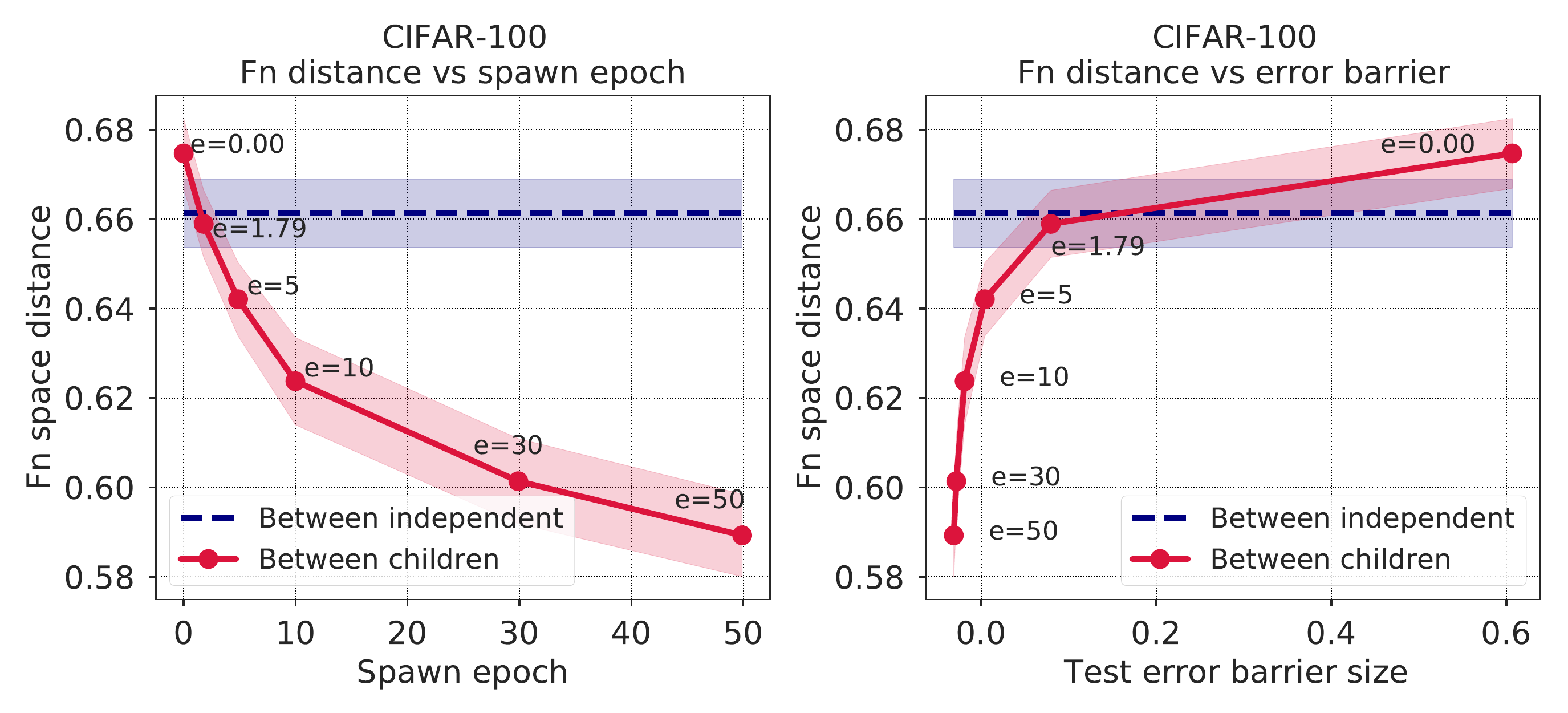}
\caption{Relation between error barrier and child function distance for ResNet20 on CIFAR 10 and 100. Left panels show how final child distance (near 200 epochs) falls off with {\it spawn} epoch (red curve). The purple baseline indicates final distance between two independent parents. Right panels plot function distance as a function of error barrier. See also \cref{fig:acc_and_fndistance_vs_epoch_resnet_cifar10and100} for detailed evolution of both quantities with training rather than spawn epoch. Error bars reflect std. dev. across the last 25 epochs. The function prediction embeddings are shown in \cref{fig:tsne_bn_main}.}
\label{fig:fnspace_distance}
\end{figure}

In \cref{fig:fnspace_distance} we explore more quantitatively the relationship between the final function space distance between children, spawn time of children, and the error barrier. This figure demonstrates that the error barrier drops to zero rapidly within 2-3 epochs, and then after that, the later two children are spawned, the closer they remain to each other.  Since these experiments were done by training parents and children at a constant learning rate over 200 epochs such that child distances stabilized, the reduction in achievable function space distance between children as a function of spawn time cannot be explained either by learning rate drops or by insufficient training time for children (see \cref{fig:acc_and_fndistance_vs_epoch_resnet_cifar10and100}). 

\section{NTK velocity slows down and stabilizes after basin fate is determined}
\label{sec:bumpvsvel}

\begin{figure}[h]
\centering
\includegraphics[width=0.8\linewidth]{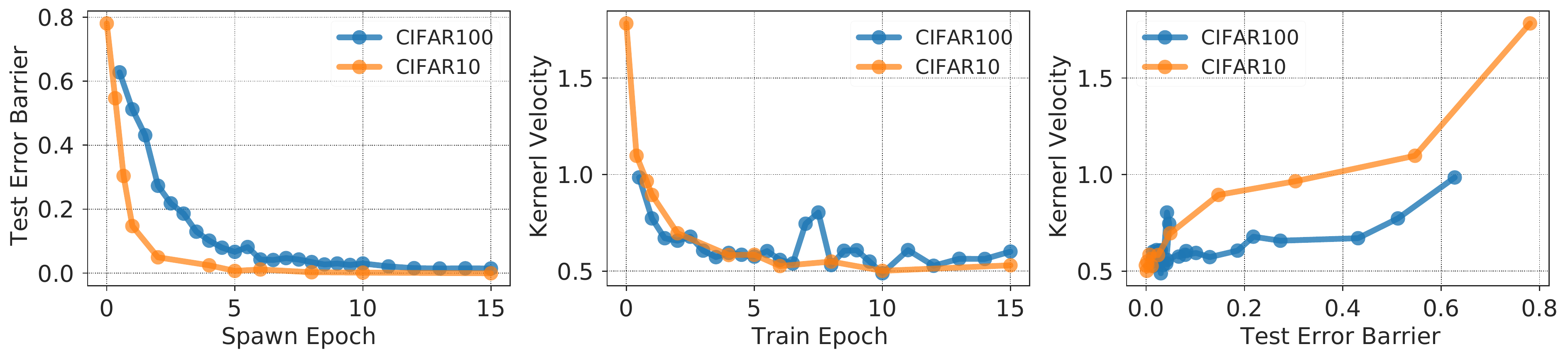}
\caption{Relation between test error barrier and kernel velocity for a ResNet20 trained on CIFAR10 and CIFAR100. Both the test error barrier between children (left) and the kernel velocity of the parent (middle) fall off and stabilize early in time and exhibit strongly correlated dynamics (right).\vspace*{-1em}}
\label{fig:kernel_distance_error_bump}
\end{figure}

We next explore the relation between error barrier and kernel velocity in \cref{fig:kernel_distance_error_bump} by zooming in on the early epochs, compared to the full training shown in \cref{fig:megaplot1}-\ref{fig:megaplot2} panels B and D. This higher resolution view clearly reveals that the early chaotic training regime is characterized by a tightly correlated reduction in both error barrier and kernel velocity, with the latter stabilizing to a low non-zero velocity after the error barrier disappears.  Thus the NTK evolves relatively rapidly until basin fate is sealed.

\section{The data-dependent NTK rapidly learns features useful for performance}
\label{sec:dataNTK}

\begin{figure}[h]
\centering
\includegraphics[width=1.0\linewidth]{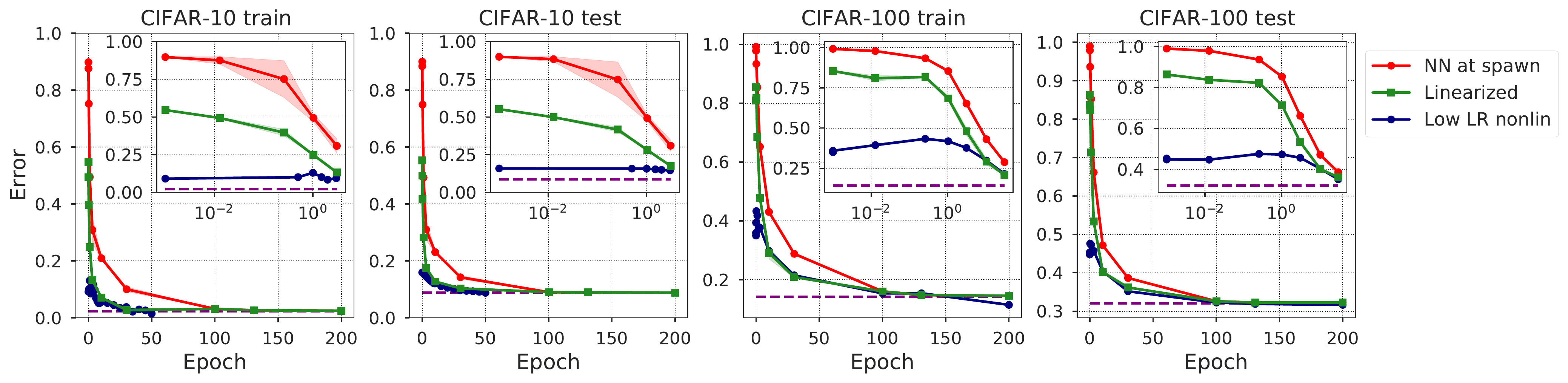}
\caption{Linearized training vs.\ ordinary training.  
The red baseline curves show the (test/train) error of a network using full nonlinear training as a function of training epoch $\tilde t$. The dashed purple constant baseline, purely for reference, indicates the error obtained at epoch $T=200$.  The green line indicates the error of a data dependent NTK obtained at training epoch $\tilde t$; i.e. the error for the green line is obtained by full nonlinear network training up to time $\tilde t$, and then subsequent linearized training with an optimal early stopping criterion for the test error. The train/test error at the optimal stopping time is plotted as a function of the {\it onset} time $\tilde t$ of linearized training, reflecting the performance of the data-dependent NTK at time $\tilde t$.   The blue curve is obtained identically to the green curve, except instead of using linearized training, we use full nonlinear training at the lowest possible learning rate after time $\tilde t$, that still ensures convergence after $1000$ epochs.  We explore the gap between the green and blue curves in \cref{fig:nonlinear_advantage}. In the Appendix in \cref{fig:taylor_WRN} we show additional results for WideResNet and in \cref{fig:taylor_adam_resnet} Taylor expansions of order 2 for ResNet.} 
\label{fig:taylorized_network_training_accuracies}
\end{figure}

The rapid evolution of the NTK during the chaotic training regime and its subsequent constant velocity motion after basin fate determination, as shown in \cref{fig:kernel_distance_error_bump}, raises a fundamental question: at what point during training does the NTK learn useful features that can yield high task performance, or even match full network training?  We answer these questions in \cref{fig:taylorized_network_training_accuracies} through a two step training protocol. We first train the full nonlinear network up to a time $\tilde t$. We then Taylor expand the full nonlinear network $f_w$ obtained at time $\tilde t$ with respect to the weights $w$, and perform linearized training thereafter up to a total time $T$. Geometrically this corresponds to training for time $\tilde t$ up to one of the intermediate green points in \cref{fig:conceptual}CD, and then subsequently training only within the green tangent space about that point in function space.  We can think of this as training with a data-dependent NTK kernel that has been learned using data from time $0$ to $\tilde t$.  Classic NTK training corresponds to a random kernel arising when the onset time $\tilde t$ of linearized training is $0$.

Using this two step procedure, \cref{fig:taylorized_network_training_accuracies} demonstrates several key findings.  First, extremely rapidly, within $\tilde t = 3$ to $4$ epochs, the data dependent NTK has learned features that allow it to achieve significantly better performance (i.e. error drops by at least a factor of 3) compared to  the classic NTK obtained at initialization (see rapid initial drop of green curves in \cref{fig:taylorized_network_training_accuracies}). Second, by about $\tilde t = 30$ to $90$ epochs, representing 15\% to 45\% of training, the data-dependent NTK essentially matches the performance of a network trained for the standard full 200 epochs (compare green curves to purple baseline in \cref{fig:taylorized_network_training_accuracies}). 
This indicates that the early chaotic training period characterized by rapid drops in error barrier and kernel velocity in \cref{fig:kernel_distance_error_bump} also corresponds to rapid kernel learning: useful information is acquired {\it within a few epochs}. This kernel learning continues, albeit more slowly after the initial chaotic learning period is over and the basin fate is already determined.

\vspace*{-0.3cm}
\section{NTK and nonlinear training remain different even at low learning rates}
\vspace*{-0.3cm}
\label{sec:nonlinadv}

\begin{figure}[h]
\centering
\includegraphics[width=0.49\linewidth]{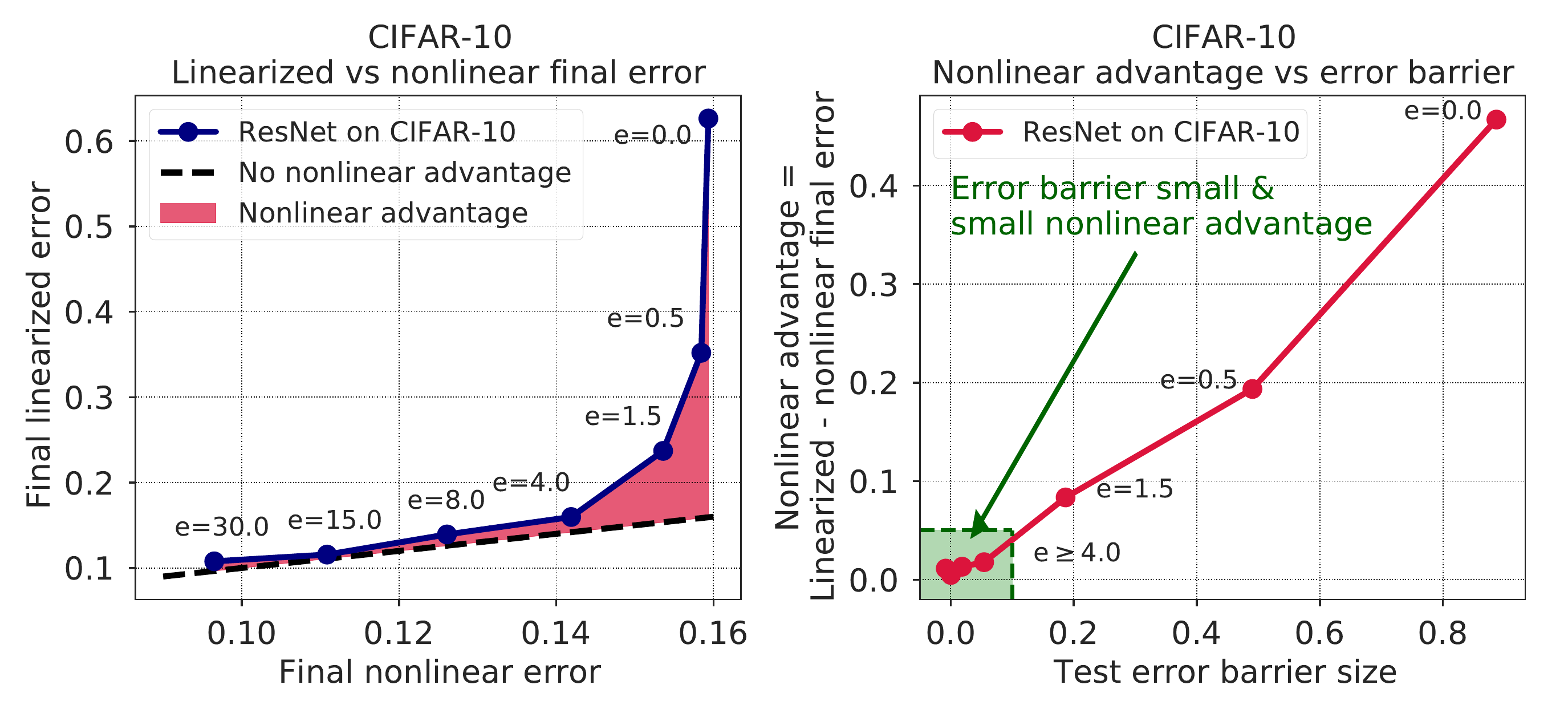}
\includegraphics[width=0.49\linewidth]{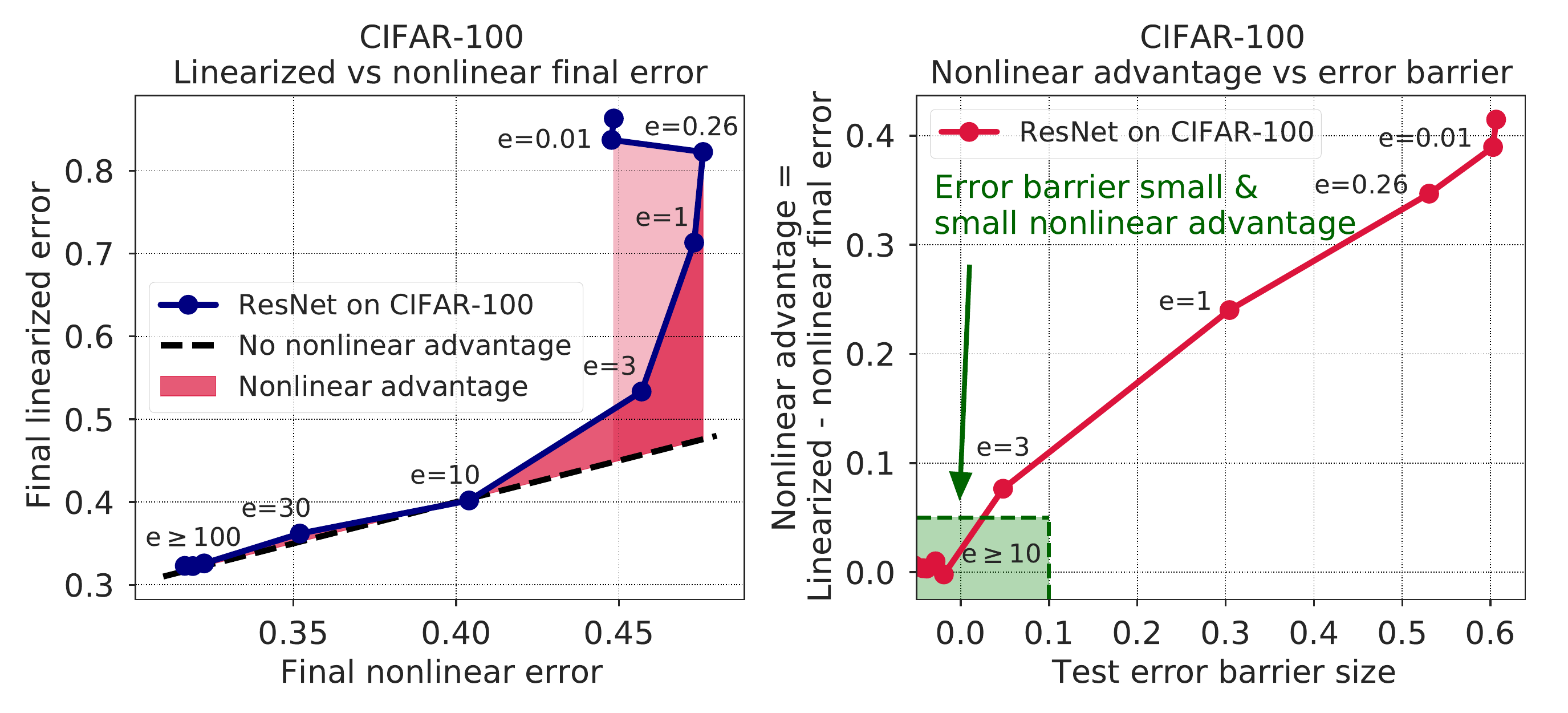}
\caption{Relation between low learning rate nonlinear error advantage and error barrier size. For each dataset, the left panel blue curves plot the error obtained by linearized training (green curve in \cref{fig:taylorized_network_training_accuracies}) against the error obtained by nonlinear low learning rate training (blue curves in \cref{fig:taylorized_network_training_accuracies}), with the epoch indicating the onset time $\tilde t$ of both. The dashed line is the unity line, so the height of the red region indicates the nonlinear advantage (error reduction due to low learning rate nonlinear training relative to linearized training). Right panels plot nonlinear advantage against error barrier size.}
\label{fig:nonlinear_advantage}
\end{figure}

In the NTK limit, which involves {\it both} infinite widths and infinitesimal  learning rates, linearized training and full nonlinear training dynamics provably coincide. However, the persistent performance gap up to 30 to 90 epochs between linearized and full nonlinear training (green curves versus purple baselines in \cref{fig:taylorized_network_training_accuracies}) indicates the NTK limit does not accurately describe training dynamics used in practice, at finite widths and large learning rates. We remove one of the two reasons for the discrepancy by comparing the same linearized training dynamics to extremely low learning rate nonlinear training dynamics (blue curves in \cref{fig:taylorized_network_training_accuracies}).  In this finite width low learning rate regime, we find, remarkably, that a significant performance gap persists between linearized and nonlinear training (the red nonlinear training advantage region in \cref{fig:nonlinear_advantage}, left), but only during the first few epochs of training, corresponding precisely to the chaotic regime before basin fate is sealed. Indeed the disappearance of this low learning rate nonlinear advantage is tightly correlated with the disappearance of the error barrier (\cref{fig:nonlinear_advantage}, right).  This indicates that while the data-dependent NTK limit can describe well the low (but not high) learning rate dynamics after the first few epochs, this same NTK limit cannot accurately describe the full nonlinear learning dynamics during the highly chaotic early phase prior to basin fate determination, {\it even when} the full nonlinear training uses very low learning rates, {\it and when} the NTK is learned from the data. We present additional experiments with Taylor expansions of order 2 on ResNet in \cref{fig:taylor_adam_resnet} and linear order for WideResNet in \cref{fig:taylor_WRN}. In \cref{app:nobn}, we also perform the same set of linearized training experiments on networks trained with no batch normalization to ensure that we observe the same effect.

\vspace*{-0.3cm}
\section{Summary of contributions and discussion}
\vspace*{-0.3cm}
\label{sec:discussion}

In summary, we have performed large scale {\it simultaneous} measurements of diverse metrics (\cref{fig:megaplot1,fig:megaplot2,fig:megaplot0}) finding a strikingly universal chaotic to stable training transition across datasets and architectures that completes within two to three epochs. During the early chaotic transient: (1) the final basin fate of a network is determined (\cref{fig:kernel_distance_error_bump} left); (2) the NTK rapidly changes at high speed (\cref{fig:kernel_distance_error_bump} middle and \cref{fig:megaplot0}D); (3) the NTK rapidly learns useful features in training data, outperforming the standard NTK at initialization by a factor of 3 within 3 to 4 epochs (\cref{fig:taylorized_network_training_accuracies} green curves); (4) even low learning rate training retains a nonlinear performance advantage over linearized NTK training with a learned kernel (\cref{fig:nonlinear_advantage} red regions); and (5) the error barrier, kernel velocity, and low learning rate nonlinear advantage all fall together in a tightly correlated manner (\cref{fig:kernel_distance_error_bump}, right) and (\cref{fig:nonlinear_advantage}, right).  After this rapid chaotic transient, training enters a more stable regime in which: (6) SGD stochasticity allows more limited child exploration in terms of function space distance, leading to smaller function diversity within basins compared to between basins (\cref{fig:children_spreading,fig:fnspace_distance}); (7) the kernel velocity stabilizes to a fixed nonzero speed (\cref{fig:kernel_distance_error_bump} middle and \cref{fig:megaplot0}D); (8) the data dependent kernel performance continues to improve, matching that of full network training by 30 to 90 epochs, of training, representing 15\% to 45\% of the full 200 epochs (\cref{fig:taylorized_network_training_accuracies} green curves). 

The empirical picture uncovered by our work is much richer than what any theory of deep learning can currently capture. In particular, the NTK theory attempts to describe the entire  nonlinear deep learning process using a {\it fixed random} kernel at initialization. While this description is provably accurate at infinite width and low learning rate, our results show it is a poor description of what occurs in practice at finite widths and large learning rates (\cref{fig:kernel_distance_error_bump,fig:taylorized_network_training_accuracies}). More interestingly, the NTK theory is even a poor description of nonlinear training at finite width and extremely low learning rates, especially during the early chaotic training phase (\cref{fig:nonlinear_advantage}). 

This rich phenomenological picture of the rapid sequential nature of the learning process could potentially yield practical dividends in terms of a theory for the rational design of learning rate schedules. For example, the timing of optimized learning rate drops coincide with the time when the data-dependent tangent kernel can achieve high accuracy. Indeed our observations are consistent with findings in \citep{leclerc2020two}. But more generally, we hope that our empirical measurements of such a rich phenomenology may serve as an inspiration for developing an equally rich unifying theory of deep learning that can simultaneously capture these diverse phenomena.

\section*{Broader Impact}
The goal of our work is to gain a better understanding of deep neural networks. This could potentially make machine learning applications more reliable and transparent in the long run.

\section*{Funding Sources}

DMR was supported, in part, by an NSERC Discovery Grant, Ontario Early Researcher Award, and a stipend provided by the Charles Simonyi Endowment. SG thanks the Simons Foundation, James S. McDonnell Foundation, NTT Research, and an NSF Career award for support. This research was in part carried out while GKD and DMR participated in the Special Year on Optimization, Statistics, and Theoretical Machine Learning at the Institute of Advanced Studies.

\section*{Acknowledgements}

The authors would like to thank Jonathan Frankle, Ekansh Sharma, and Mufan Li for feedback on drafts, and Shems Saleh for helping to produce \cref{fig:conceptual}.

\small

\printbibliography

\clearpage
\appendix

\section{Function distance between children runs}
\begin{figure}[b!]
\centering
\includegraphics[width=0.49\linewidth]{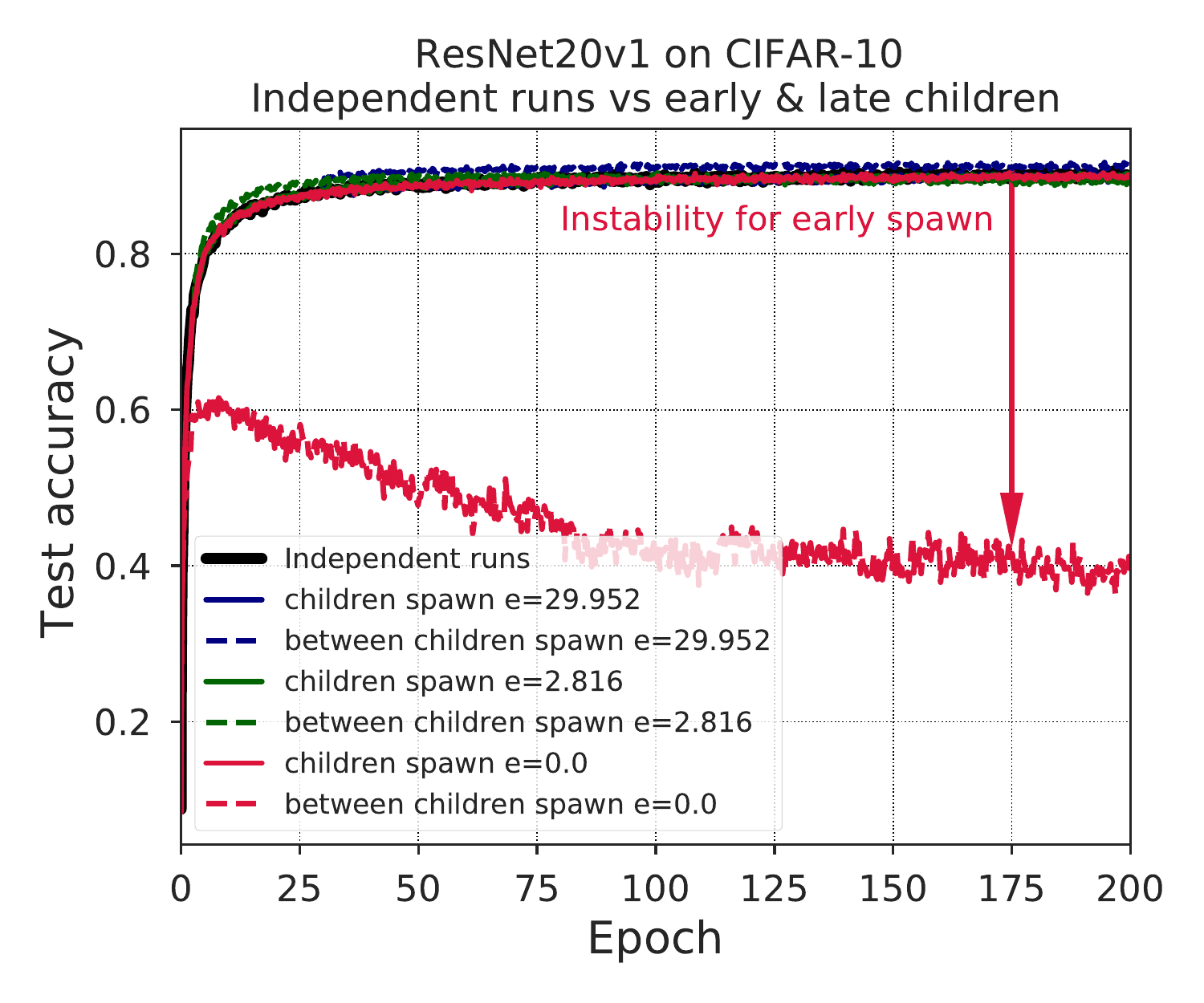}
\includegraphics[width=0.49\linewidth]{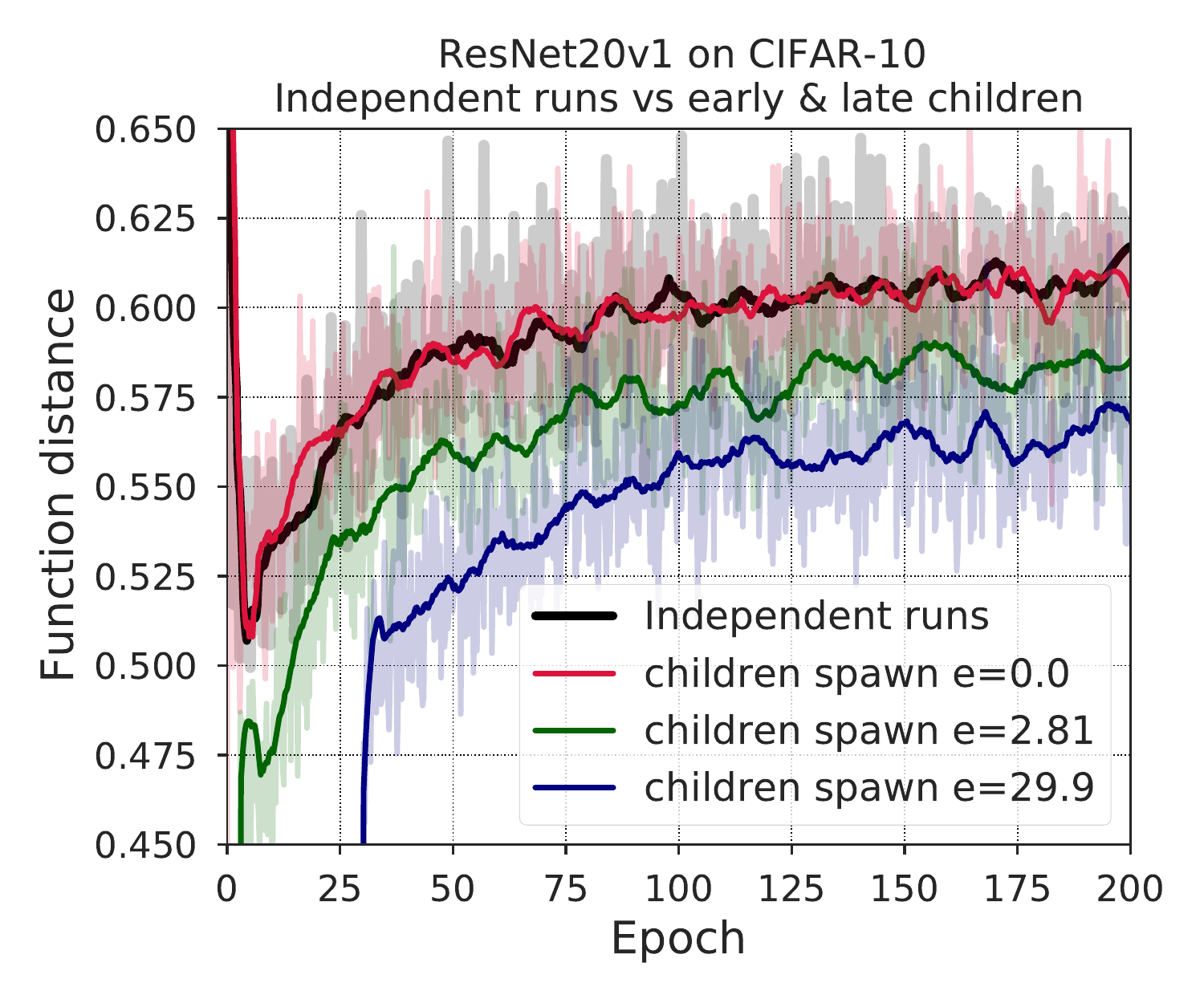} \\
\includegraphics[width=0.49\linewidth]{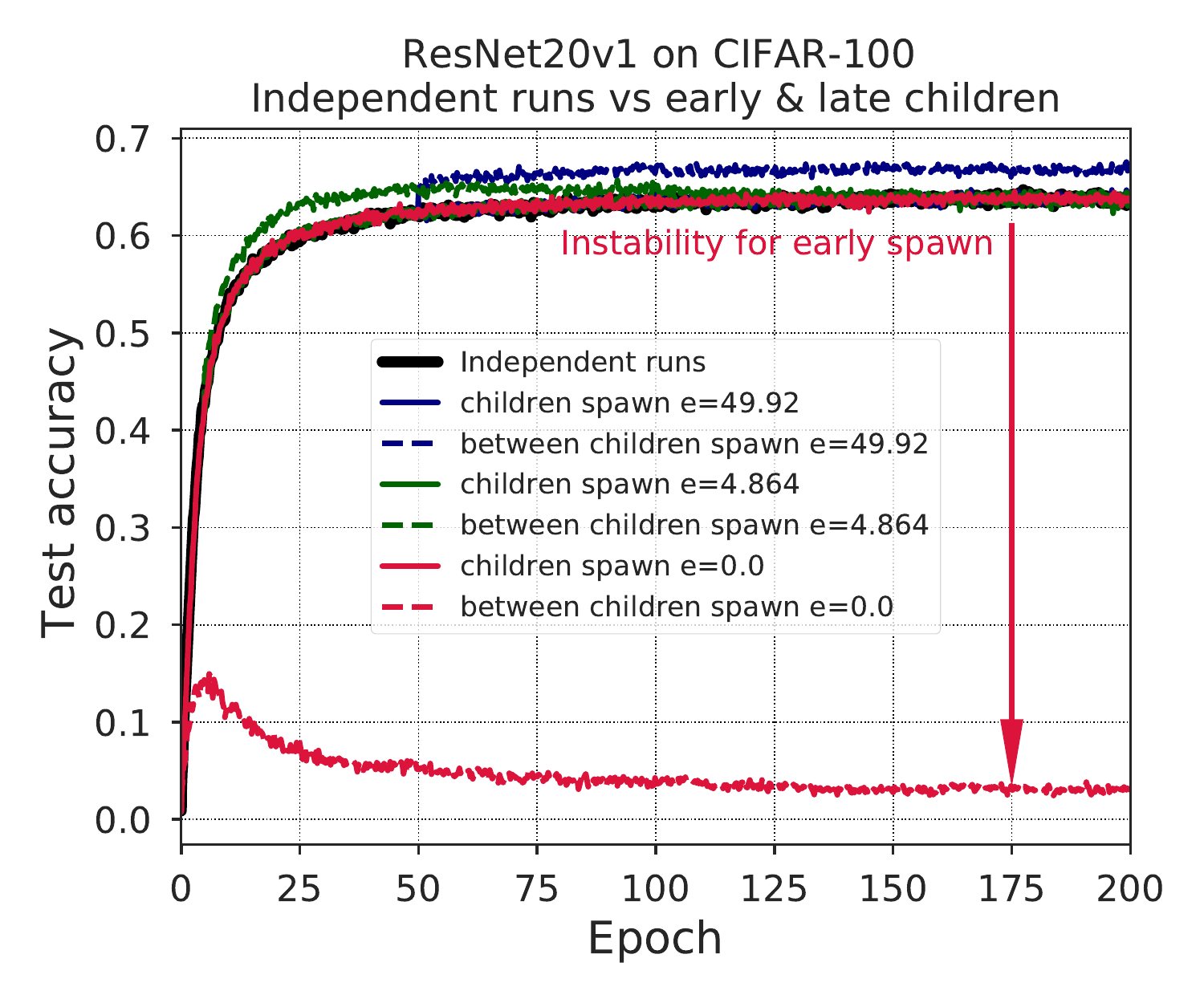}
\includegraphics[width=0.49\linewidth]{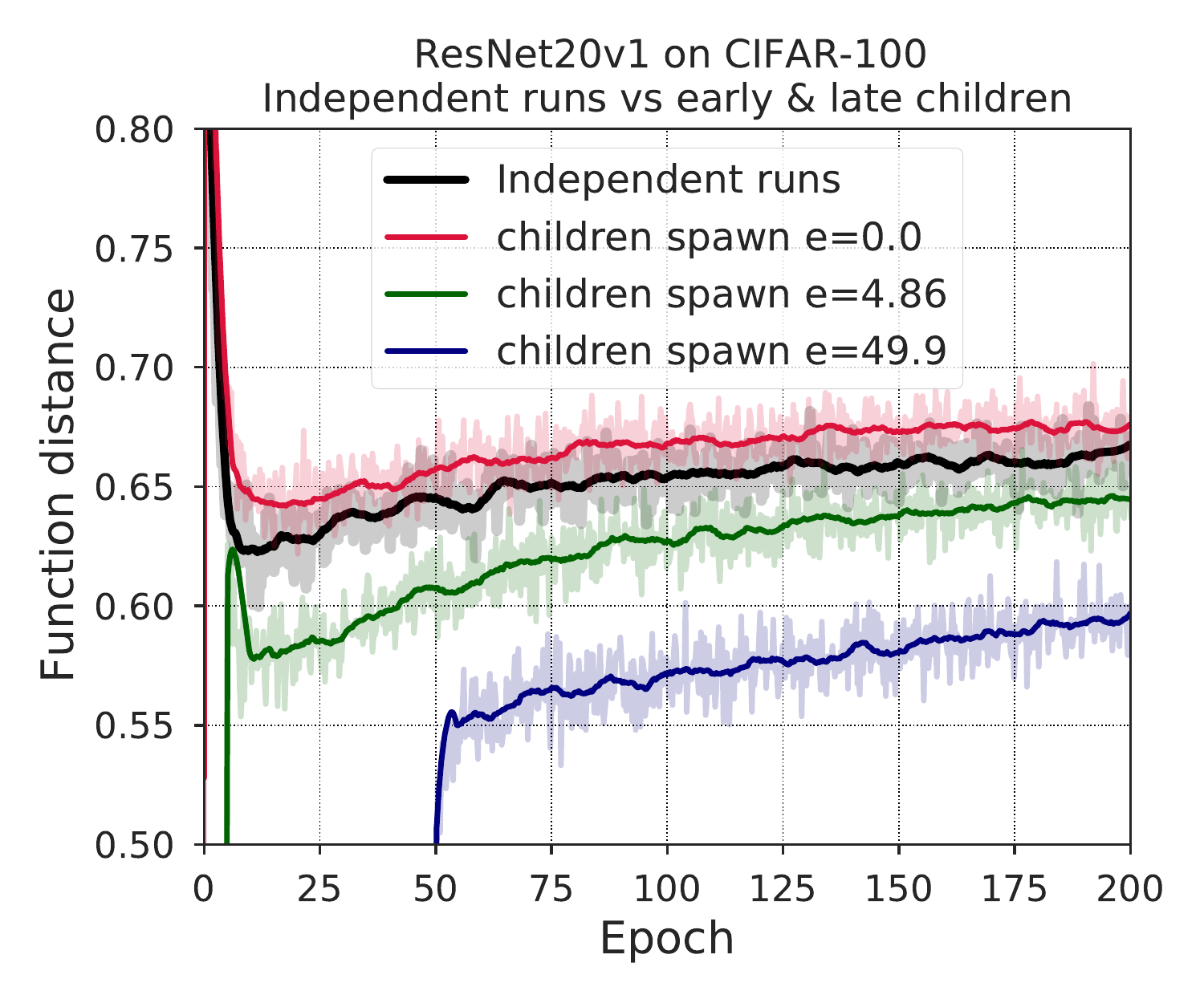}
\caption{The test accuracy and function space distance between independent runs and children spawn at different epochs for ResNet20 on CIFAR-10 (top row) and CIFAR-100 (bottom row). The noisy curves show the raw data recorded every $1/3$ of an epoch, while the thick lines show the moving average of $\pm 5$ epochs.}
\label{fig:acc_and_fndistance_vs_epoch_resnet_cifar10and100}
\end{figure}

The function distance between the children runs is shown in \cref{fig:acc_and_fndistance_vs_epoch_resnet_cifar10and100} (right column) for ResNet20 on CIFAR-10 (top row) and CIFAR-100 (bottom row). The signal is relatively noisy from iteration to iteration, we therefore overlay the raw data with a smoothed out version with a window of $\pm5$ epochs in \cref{fig:acc_and_fndistance_vs_epoch_resnet_cifar10and100}. 

\begin{figure}[t!]
\centering
\includegraphics[width=1\linewidth]{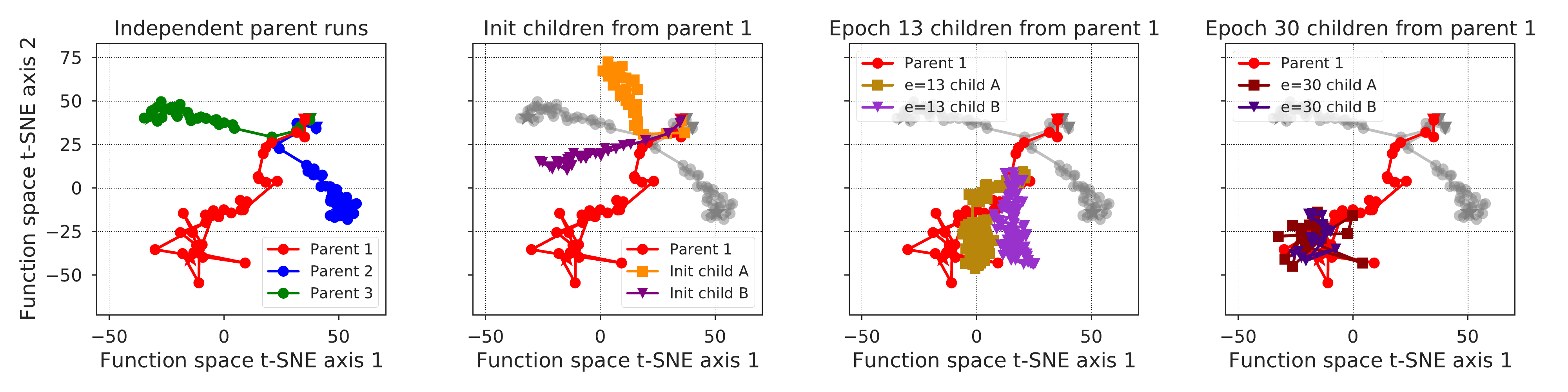}
\caption{A visualization of the function space motion during training of $3$ parents and $3$ pairs of children spawned at early, late and intermediate times in a common t-SNE embedding. Different networks are highlighted in different plots from left to right. Siblings spawned later remain closer to each other and to their parent at the end of training.}
\label{fig:tsne_bn_main}
\end{figure}

We also produced a t-SNE \citep{vanDerMaaten2008} visualization of parent and children evolution in the function space. To do that, we took predictions of the parent and children runs at different stages of their training on the test set. We then flattened the vector of predicted probabilities for all images and all of their classes into a single long vector, one for each stage of training of a network. We then used the t-SNE embedding to embed all parent and children runs into a 2D space. For the individual panels in 
\cref{fig:tsne_bn_main}, we highlighted the relevant embedded points, however, due to the nature of t-SNE all predictions had in fact been embedded together.

\section{Definitions of additional metrics}
\subsection{Logit gradient centroid alignment}
\label{sec:logitgradcentroids}
The logit gradient centroid for each class $k$ is defined as 
$\mu_w^k = \frac 1 m \sum_{i=1}^m \grad_w (f_w(x_i))_k$. Previous work \citep{fort2019emergent,papyan2019measurements} has shown that the span of the $K$ logit gradient centroids approximately tracks an important local quantity: the span of the top $K$ directions of maximal Hessian curvature (see also see \cref{centroids_high_curvature}).
Thus the span of the logit gradient centroids describes the orientation of the walls of the basins shown schematically in \cref{fig:conceptual}AB, and two trained networks with highly dissimilar logit gradient centroids likely lie in differently oriented basins. 
In order to evaluate how logit gradient centroids compare at $w$ and $w'$, we compute average cosine similarity 
\[\label{eq:cossim}
 \frac 1 K \sum_{k=1}^K {\mu_w^k \cdot \mu_{w'}^k } / {\big(\norm{\mu_w^k } \norm{\mu_{w'}^k }\big)},
 \] 
which we refer to as logit gradient centroid alignment.

\subsection{Logit gradient centroids and the top Hessian eigenvectors}
\label{centroids_high_curvature}

Here we note qualitative relations between the space of logit gradient centroids and the principal Hessian subspace. 
First, without loss of generality, assume $K=1$ and consider a single data point $(x,y)$.
Consider mean squared error (MSE) loss, with empirical risk term $\Risk{S}{w} = (f_w(x) - y)^2$. Then
\[
\frac{\dee^2 \Risk{S}{w}}{ \dee w_i \dee w_j} = \frac{\dee f}{\dee w_i} \frac{\dee f}{\dee w_j} + \Big[  (f_w(x)-y) (H_w)_{i,j} \Big] ,
\]
where 
$H_w = \grad^2 f_w(x)$ (as defined in \cref{sec:preliminaries}).

If $f_w = f_{w_0} + \ip{\grad f_{w_0}}{w-w_0}$, then $H_w$ is zero and the second term vanishes. 
Similarly, for $K \in \Nats$ and a training set $S$ of size $m$,  the Hessian of MSE error loss is equal to $\grad^2 \Risk{S}{w} = \frac{1}{K^2 m} \sum_{k,k',x\in S_x}  (J_{w_0}(x))_k (J_{w_0}(x))_{k'}^T$.

Since $(J_w(x))_k = \grad_w (f_w(x))_k $, and logit gradient centroids are defined as
\[
\mu_w^k = \frac 1 m \sum_{i=1}^m \grad_w (f_w(x_i))_k,
\]
we decompose $(J_w(x))_k = \mu_w^k + e_w^{n,k}$, for $w=w_0$.
This decomposition of $J_w$  has been previously studied in \citep{fort2019emergent,papyan2019measurements}.
Then $\grad^2 \Risk{S}{w} = \frac{1}{K^2 m}  \sum_{k,k'} \mu_w^k (\mu_w^{k'})^T + \mathcal{O}(e) $.

According to our empirical results as well as other literature \citep{Papyan2019MeasurementsOT,fort2019emergent}, the logit gradient centroids  are mutually almost orthogonal. 
Therefore, $\grad^2 \Risk{S}{w} \approx \sum_k \| \mu_w^k \|^2$. 
For mutually orthogonal gradient centroids $\mu_w^k$, this amounts to a singular value decomposition with $K$ non-zero singular values $\| \mu_w^k \|^2$ associated with singular vectors $\mu_w^k$. 

While the relative length of $\mu_w^k$ can be changing with training time, the empirically observed stability of their directions $\mu_w^k$ makes
the Hessian eigenvector associated with the highest eigenvalue constrained to lie primarily within the vector space defined by the span of $\{ \mu_w^k \}_{k \in \{1,...,K\}}$. 
This subspace of dimension $K$ ($K=10$ for CIFAR-10, $K=100$ for CIFAR-100) has a significantly lower dimensions than the typical weight space of the network. 

When evaluating the cosine similarity of the logit gradient centroids as in \cref{eq:cossim}, we are approximately estimating the overlap between the low dimensional subspaces to which the sharpest directions of the Hessian are constrained between the two networks.

\subsection{Escape threshold}

\newcommand{\sobjective}[1]{\Risk{S}{#1}}
\newcommand{\hessianloss}[1]{L_{#1}}
\label{app:escape threshold}
Consider iterative optimizers that at each iteration perform an update on the weights of the form
\[
w_{t+1} = w_t - \eta \Delta_t(\minibatch),
\]
where $\minibatch$ is a minibatch, i.e., a random subset $\minibatch \subset S$.

Let $\hessianloss{t} = \nabla^2 \sobjective{w_{t}}$ be the Hessian of the empirical loss.
Then, by a second order Taylor expansion of $\Risk{S}{w_{t+1}}$ around $w_t$, we have
\[
\label{eq:taylor1}
 \sobjective{w_{t+1}}  - \sobjective{w_{t}}
     \approx  
     \frac 1 2 \eta^2 \tuple{\Delta_{t}(\minibatch),\hessianloss{t} \Delta_{t}(\minibatch) } - \eta \tuple{ \Delta_{t} (\minibatch), \nabla \sobjective{w_{t}} } .
\]

The loss after one iteration decreases if the difference $ \sobjective{w_{t+1}}  - \sobjective{w_{t}} $ is negative. Under the second order approximation, the condition for non-increasing loss is equivalent to
\[\label{eq:escapecondition}
 \eta \frac{1}{\| \grad \Risk{S}{w_t} \|^2 } \Big( \ip{ \Delta_t (\minibatch)}{\hessianloss{t} \Delta_t(\minibatch)} - 2 \ip{\Delta_t(\minibatch)}{\nabla \Risk{S}{w_t}}  \Big) \leq 0.
\]

We refer to left hand side term in \cref{eq:escapecondition} as the \emph{escape threshold}.  If the escape threshold is below zero, then the trajectory will be descending in the quadratic basin, under the assumption that the local quadratic approximation is accurate. If the escape threshold is positive, the loss will increase or the trajectory will escape the quadratic basin.

Gradient descent (GD) update is $\Delta_t(S) = \grad \Risk{S}{W_t}$. Combining this with the bound
$\tuple{\Delta_t,\hessianloss{t} \Delta_t} \leq \lambda_t \| \Delta_t \|^2 $, where $\lambda_t$ is the spectral norm of $\hessianloss{t}$, \cref{eq:escapecondition} simplifies to $2 - \eta \lambda_t$. We refer to this term
as the escape threshold for GD.

\section{Additional results}
\label{app:additionalfigs}

Here we present additional experiments similar to the ones in \cref{fig:megaplot0,fig:megaplot1,fig:megaplot2} that include measurements of logit gradient centroid clustering, Hessian spectral norms and escape-time threshold analysis (for the networks with no batch norm).

\subsection{Diverse metrics for loss landscape and training are highly correlated}
\label{sec:corrmetrics}
\vspace{-0.8em}
We compute all of above metrics for SOTA networks in \cref{fig:megaplot1,fig:megaplot2,fig:megaplot3,fig:megaplot4,fig:megaplot5}
(see \cref{app:expdetails} for details of training and hyperparameters).
The top rows of \cref{fig:megaplot3,fig:megaplot4,fig:megaplot5} describe the dynamics of parent training.  
From left to right, the test and training error drop, as well as the top Hessian eigenvalue drops over time, and all three distances (ReLU pattern, mean logit gradient, and kernel distance) computed on a parent run between pairs of training epochs reveal 
a rapid change around a very early point at about 5 epochs, followed by slow freezing of all three quantities.  
In particular, there is significant kernel learning.  Moreover, there is a period of time early on where the top Hessian eigenvalue $\lambda > 2/\eta$ where $\eta$ is the learning rate. This large learning rate condition would be necessary for gradient descent to escape a quadratic minimum (\cref{fig:conceptual}E, top).

The bottom row computes various distances between pairs of children at the \emph{end} of training as a function of the common time $t_s$ at which the pairs of children were spawned.  
These plots indicate that at an \emph{extremely} early spawn epoch (x-axis) of around $t_s=1$, the basin fate of the two children is sealed by their parent. 
Beyond $t_s=1$, the two children end up in the same basin, as evidenced by the lack of a loss barrier along a linear path between them, and are much closer to each other, as measured by distances in (from left to right), weight space, ReLU pattern, logit gradient, function space, and kernel space. 
In contrast, before this early spawn epoch of $1$, the final basin choice of the children displays a chaotic sensitivity to SGD  steps, as evidenced by a large loss barrier and larger distances. 
Intriguingly, significant, albeit slow kernel learning continues after basin fate selection.   

\subsection{Further discussion}

\begin{figure}[t!]
    \centering
    \includegraphics[width=\linewidth]{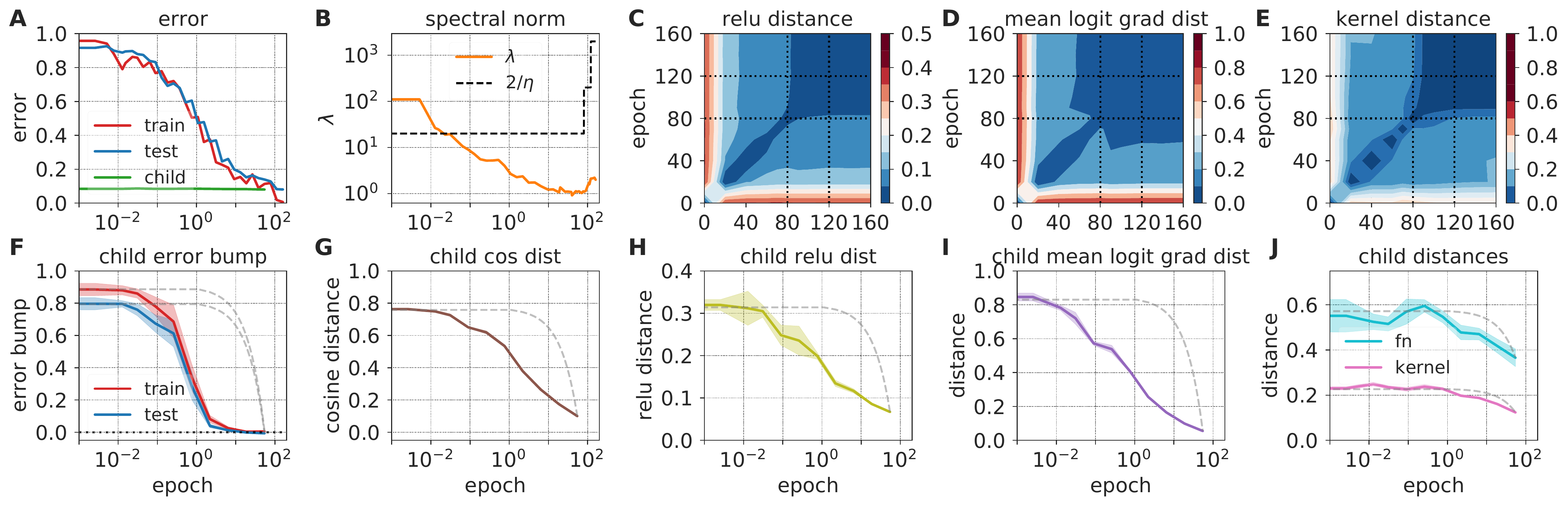}
    \caption{ResNet20 with batchnorm trained on CIFAR10 using momentum.
    }
    \label{fig:megaplot7}
    \centering
    \includegraphics[width=\linewidth]{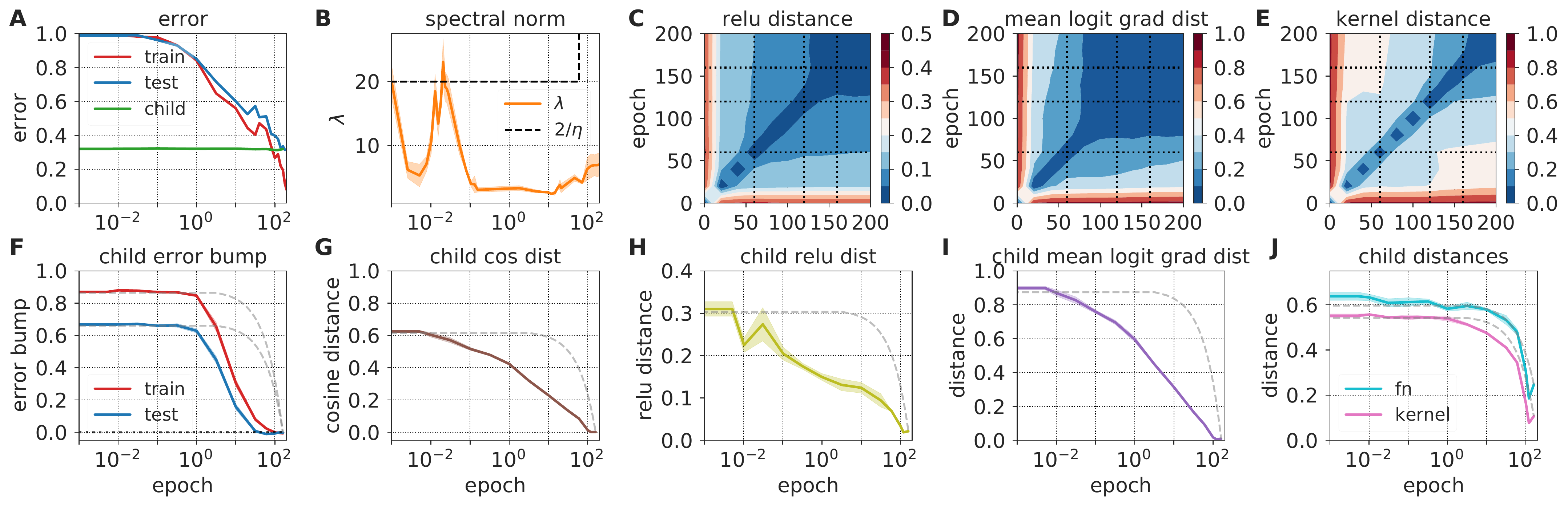}
    \caption{ResNet20 with batchnorm trained on CIFAR100 using momentum.
    }
    \label{fig:megaplot8}
    \centering
    \includegraphics[width=\linewidth]{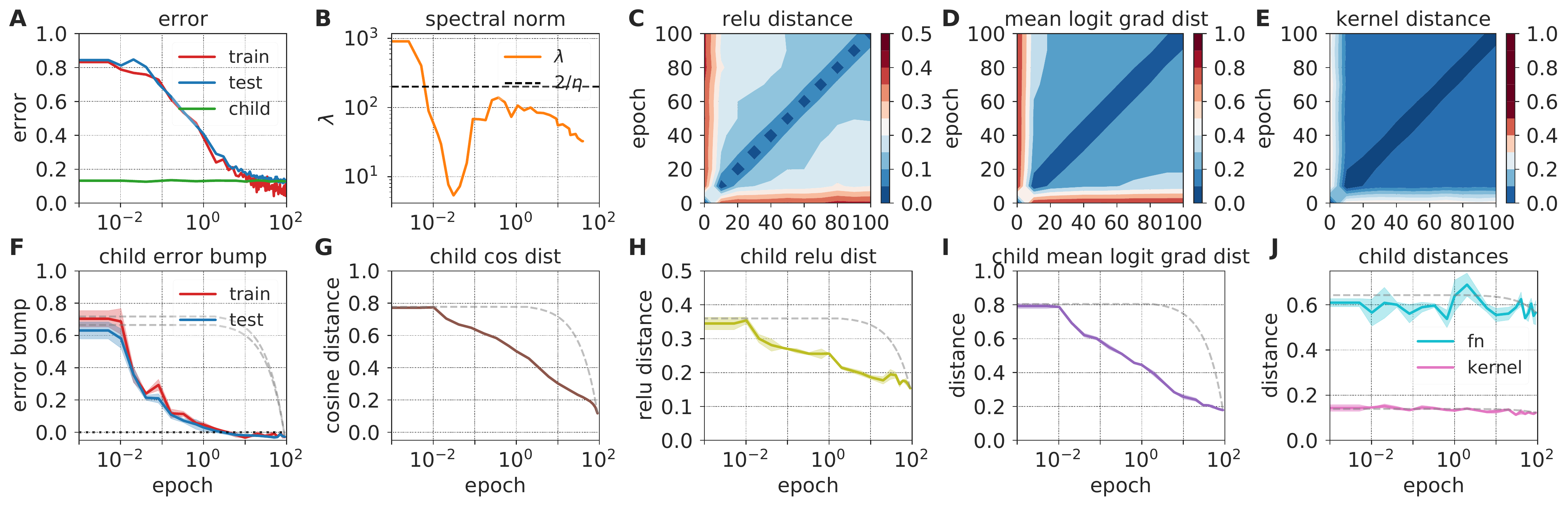}
    \caption{SimpleCNN trained on CIFAR10 using Adam. \newline
    Grey dashed lines in F--J represent a straight line between the y axis values at epoch 0 and the final epoch. X-axis in F--J indicates the spawning epoch.
    Dashed lines in C--E mark epochs when the learning rate was dropped.
    In A, `child` line represents the final test error of a child spawn at epoch indicated on the x-axis. \cref{fig:megaplot7}B, \cref{fig:megaplot8}B, and \cref{fig:megaplot3}B depicts the spectral norm of the Hessian compared to the learning rate $\eta$. \emph{These are additional results extending \cref{fig:megaplot1,fig:megaplot2,fig:megaplot0}}
    }
    \label{fig:megaplot3}
\end{figure}

\begin{figure}[t!]
    \centering
    \includegraphics[width=\linewidth]{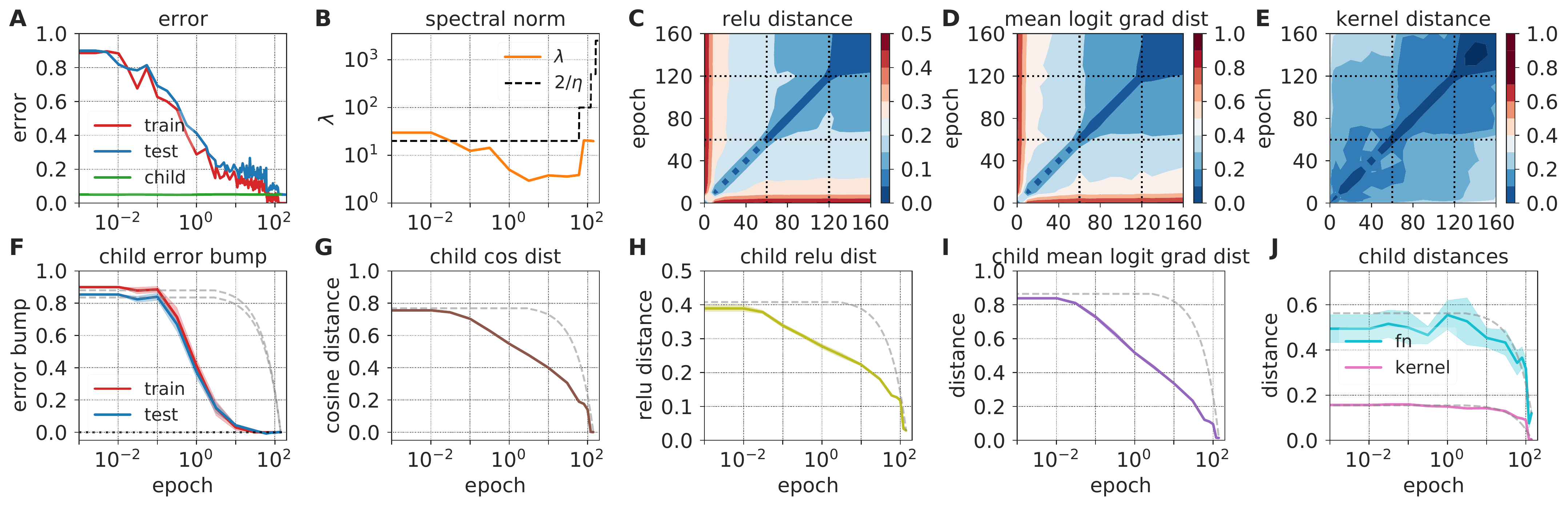}
    \caption{WideResNet-16-4 trained on CIFAR10 using momentum. 
    }
    \label{fig:megaplot4}
    \centering
    \includegraphics[width=\linewidth]{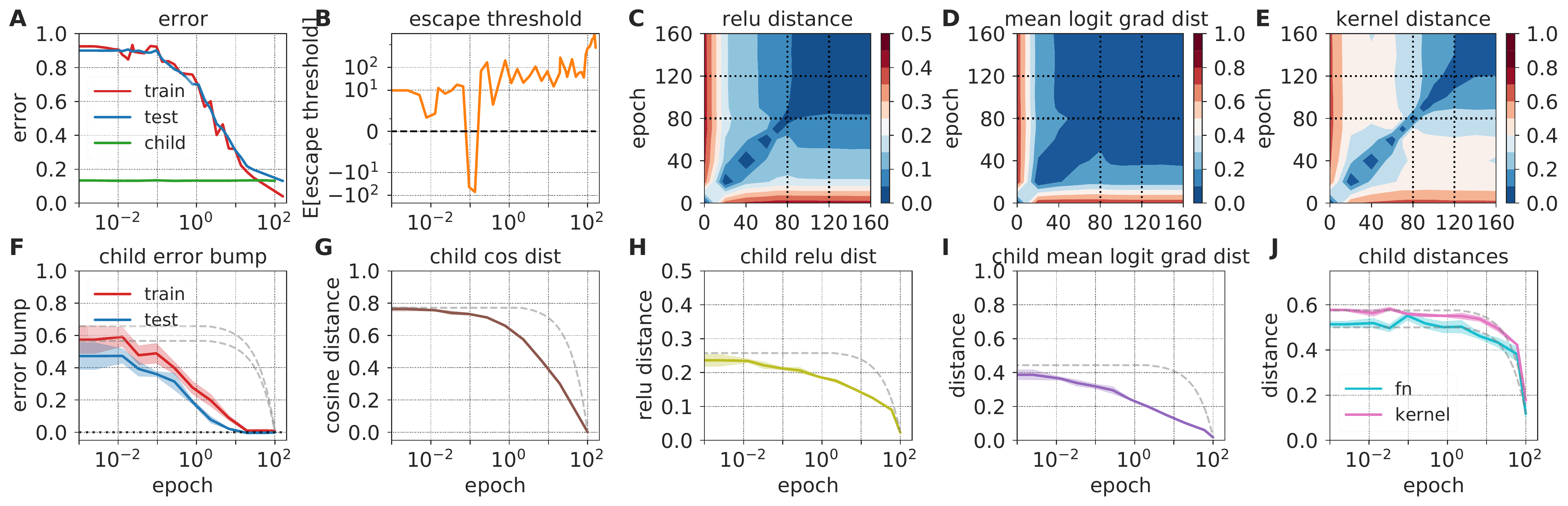}
    \caption{ResNet20 without batchnorm trained on CIFAR10 using Adam. 
    }
    \label{fig:megaplot5}
    \centering
    \includegraphics[width=\linewidth]{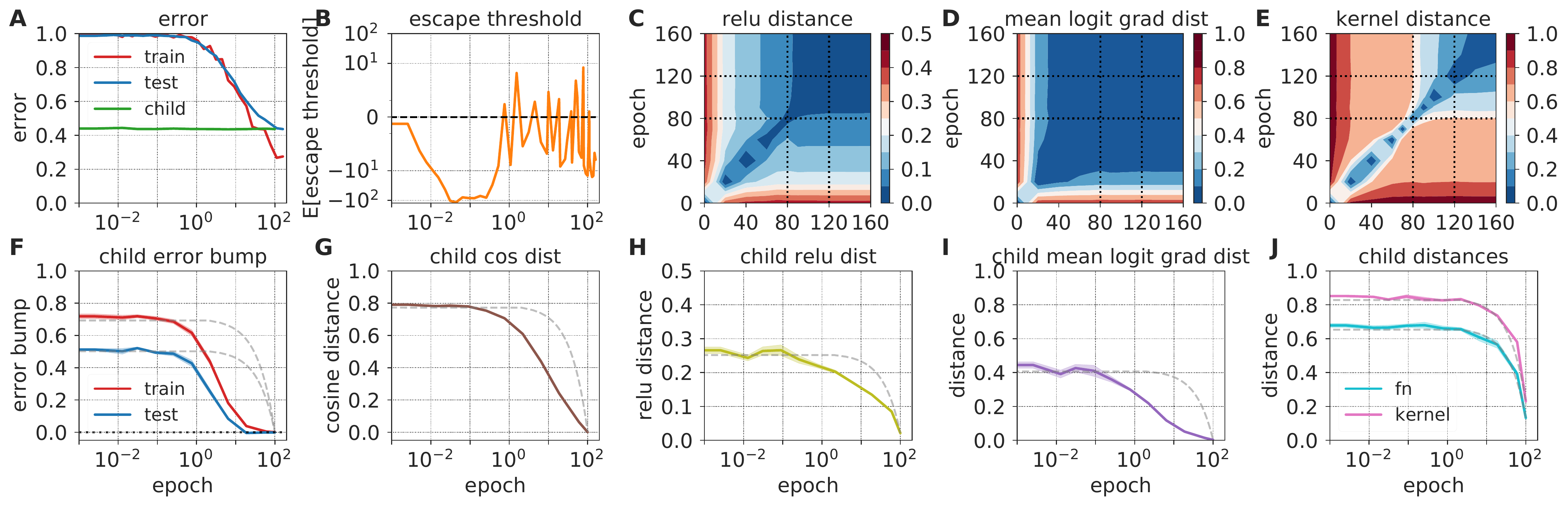}
    \caption{ResNet20 without batchnorm trained on CIFAR100 using Adam. \newline Grey dashed lines in F--J represent a straight line between the y axis values at epoch 0 and the final epoch. X-axis in F--J indicates the spawning epoch.
    Dashed lines in C--E mark epochs when the learning rate was dropped.
    In A, `child` line represents the final test error of a child spawned at epoch indicated on the x-axis. \cref{fig:megaplot4}B depicts the spectral norm of the Hessian compared to the learning rate $\eta$. \cref{fig:megaplot5}B and \cref{fig:megaplot6}B look at the escape threshold (See \cref{app:escape threshold}).
    \emph{These are additional results extending \cref{fig:megaplot1,fig:megaplot2,fig:megaplot0,fig:megaplot7,fig:megaplot8,fig:megaplot3}}
    }
    \label{fig:megaplot6}
\end{figure}

The integrative analysis of diverse measurements during neural network training (\cref{fig:megaplot1,fig:megaplot2,fig:megaplot3,fig:megaplot4,fig:megaplot5,fig:megaplot6}) and the results of linearized training (\cref{fig:taylorized_network_training_accuracies}) reveal a uniform and striking story. 
First, very early in the training process---between 1 and 10\%---of training time, the final basin fate of the neural network is determined. 
After this point, large scale motion of the neural network is no longer influenced by SGD noise and the networks trained with different SGD noise converge to low loss points in the same (linearly connected) basin. 
This is indicated by the fact that children spawned after this point are linearly connected through low error networks: the error barrier between spawned children goes below 0 at around 10 epochs for all networks. 
Additionally, the escape threshold and spectral norm analysis reveals that, after this point, the learning rate is small enough to keep the network within the local quadratic approximation of the loss surface, further supporting the hypothesis that the network does not leave the basin. 
Once the network is in the basin, we find that various metrics of network distance start decreasing. 

\begin{figure}[ht]
\centering

\includegraphics[width=1.0\linewidth]{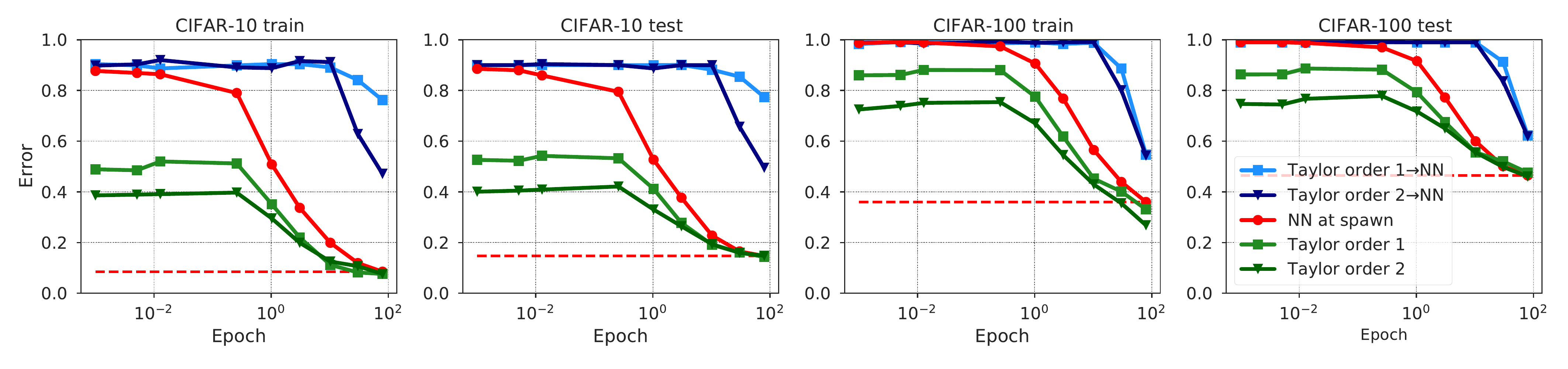}
\caption{Taylorized training versus ordinary training for ResNet20 without batch norm and trained with Adam on CIFAR-10 and CIFAR-100. Compared to \cref{fig:taylorized_network_training_accuracies}, we included the second order Taylor expansions of the neural network as well as the first order. While the second order leads to lower error whenever we spawn in along the training trajectory, it still preserves the characteristic shape of the curve for the first order, where the training of the Taylorized model only starts getting better at a particular epoch. The dashed red line shows the performance of the neural network at epoch 100 which is not its peak performance yet, since it would still improve past this point.}
\label{fig:taylor_adam_resnet}

\includegraphics[width=1.0\linewidth]{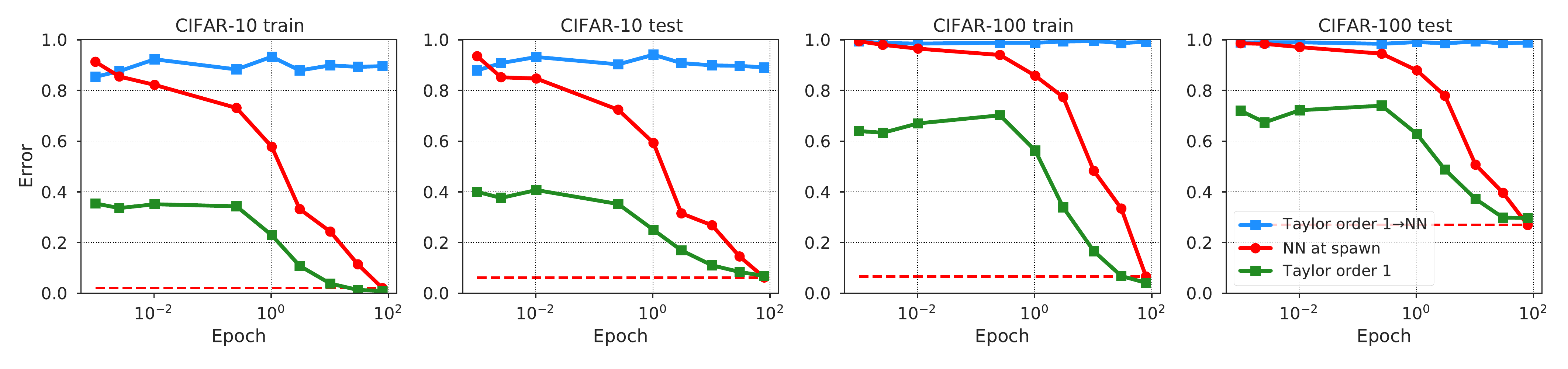}
\caption{Linearized training versus ordinary training for WideResNet on CIFAR-10 and CIFAR-100. Similarly to \cref{fig:taylorized_network_training_accuracies}, we observe that the error of the trained linearized model retains the same shape, where it only starts getting better after a particular epoch. In contrast, the accuracy of the original model with the trained linearized model's weights plugged back in does not improve at all, suggesting that the WideResNet low loss basins might not be geometrically captured as easily by the linearized approximation. The dashed red line shows the performance of the neural network at epoch 100 which is not its peak performance yet, since it would still improve past this point.}
\label{fig:taylor_WRN}

\end{figure}

Our linearized training analysis (\cref{fig:taylor_adam_resnet,fig:taylor_WRN}) reveals further interesting features about this early period of training. 
Very early in the training process, before an epoch is completed, the data-dependent NTK---the first order approximation of the neural network---rapidly starts learning useful information about the data as evidenced by the fall off of the green lines in \cref{fig:taylorized_network_training_accuracies,fig:taylor_adam_resnet,fig:taylor_WRN}. 
This occurs uniformly across networks and datasets. 
When a kernel machine is trained with these data dependent features, it performs significantly better than the NTK at random initialization. 
In fact, less than halfway through training, the data-dependent neural tangent kernel machine performs nearly as well as as the full network. 

These observations have two important implications. 
First, while training the NTK at random initialization may not be very representative of training finite sized networks, across a range of networks, the data-dependent NTK obtained from a small amount of training is. 
Second, the features built up by the NTK relatively early in the training process are sufficient for achieving low errors competitive with the full non-linear networks.

While deep learning research often focuses on improving accuracy towards the end of training, our results show that the early phase of training is important for determining the final fate of the network. 
A better understanding of this phase may provide us with tools to diagnose and improve networks early on in the training process, thus decreasing the cost of training neural networks.


\subsection{Nonlinear advantage and function space distance without batch norm}
\label{app:nobn}
\begin{figure}[h]
\centering
\includegraphics[width=0.50\linewidth]{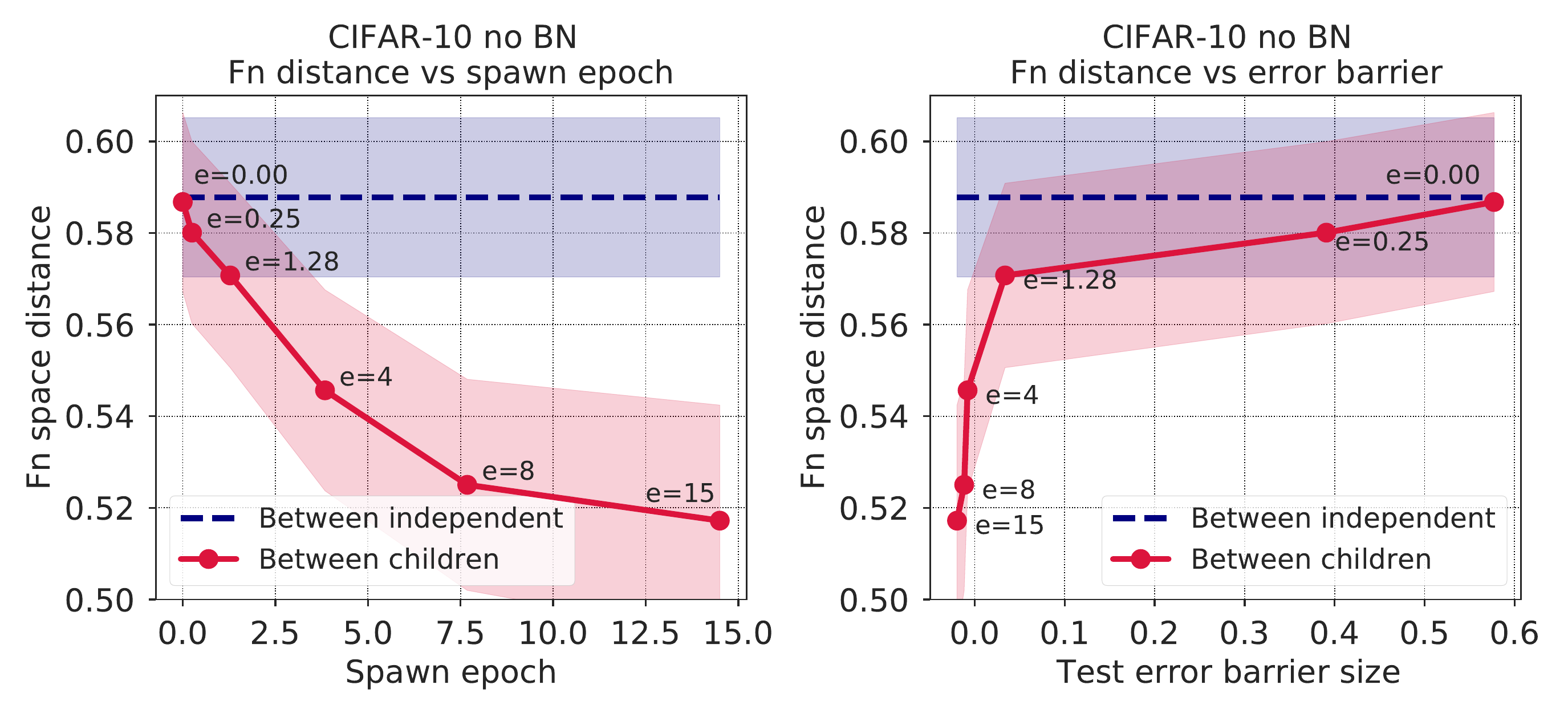}
\caption{Relation between error barrier and child function distance for ResNet20 \textit{without} batch normalization on CIFAR-10. Left panel shows how final child distance (near 200 epochs) falls off with {\it spawn} epoch (red curve). The purple baseline indicates final distance between two independent parents. Right panels plot function distance as a function of error barrier. Error bars reflect std. dev. across the last 5 epochs.}
\label{fig:fnspace_distance_noBN}
\end{figure}

\begin{figure}[h]
\centering
\includegraphics[width=0.50\linewidth]{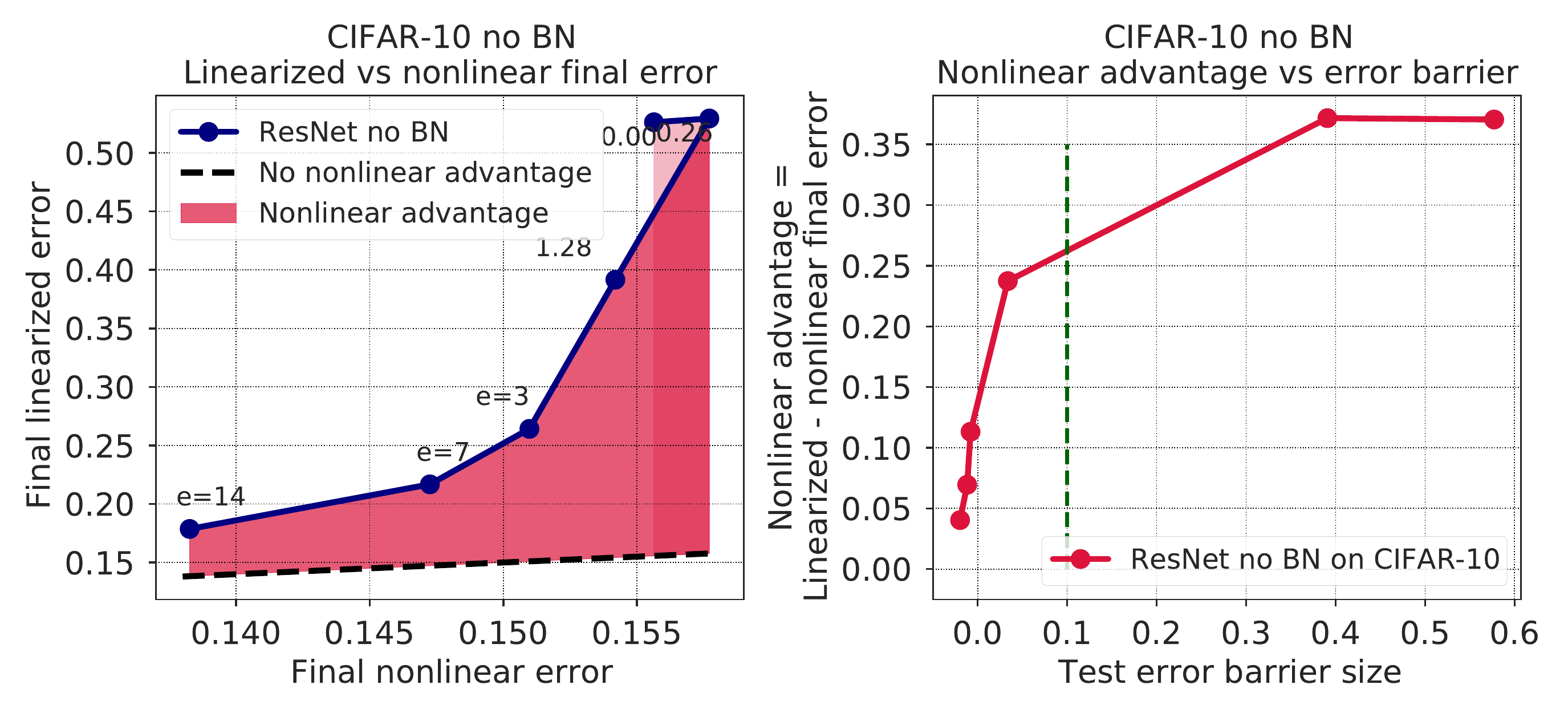}
\caption{Relation between low learning rate nonlinear error advantage and error barrier size for ResNet20v1 \textit{without} batch normalization. On the left panel blue curves plot the error obtained by linearized training (green curve in \cref{fig:taylorized_network_training_accuracies}) against the error obtained by nonlinear low learning rate training (blue curves in \cref{fig:taylorized_network_training_accuracies}), with the epoch indicating the onset time $\tilde t$ of both. The dashed line is the unity line, and so the height of the red region indicates the nonlinear advantage, or error reduction obtained by low learning rate nonlinear training relative to linearized training. Right panels plot this nonlinear advantage against error barrier size.}
\label{fig:nonlinear_advantage_noBN}
\end{figure}

In \cref{fig:fnspace_distance} we show the function space distance between children runs and its correlation with the error barrier for a network with batch normalization. The nonlinear advantage with batch normalization is also shown in \cref{fig:nonlinear_advantage}. While state-of-the-art models use batch normalization and we therefore focused on it in the main text, we wanted to confirm that the same phenomena are observed for networks without batch normalization. \cref{fig:fnspace_distance_noBN} shows the function space results, and \cref{fig:nonlinear_advantage_noBN} shows the nonlinear advantage for ResNet20v1 \textit{without} batch normalization trained on CIFAR-10 with Adam at learning rate $10^{-3}$.

\section{Experimental details}
\label{app:expdetails}
\subsection{Networks}

\textbf{SimpleCNN:} 
SimpleCNN is a 6 layer fully convolutional neural network. 
Each convolution has a 3x3 kernel, stride 1 and bias. 
The layers have 32, 32, 64, 64, 128, and 128 channels from the first to the last layer. 
The weights are initialized using Kaiming initialization \citep{he2015delving} and the biases are initialized to 0.
There is a 2D maxpooling with a 2x2 kernel and stride 2 after layers 2, 4, and 6. 
Layer 6 is followed by a 2d global average pool which results in a 1x128 unit feature vector from which the classes are linearly predicted.

\textbf{ResNet20:}
We use the ResNet20 used with CIFAR-10/100 data in the original paper \citep{he2015deep}.

\textbf{ResNet20 without batchnorm (RN no BN):}
Same as ResNet20 but with batchnorm turned off.

\textbf{WideResNet-16-4 (WRN-16-4):}
We use the WideResNet-16-4 described in \citep{zagoruyko2016wide} (16 layers, widen factor 4, no dropout).

\subsection{Training Details}

See \cref{tab:training-deets} for training details and hyperparameters.
\begin{table}[]
\begin{tabular}{|c|c|c|c|c|c|c|c|c|}
\hline
Network    & Dataset  & Opt  & LR   & Mom & WD   & LR Decay & Decay Epochs & Total Epochs \\ \hline
Resnet20   & CIFAR10  & SGD  & 1e-1 & 0.9 & 1e-4 & 0.1      & 80, 120      & 160          \\
Resnet20   & CIFAR100 & SGD  & 1e-1 & 0.9 & 1e-4 & 0.1      & 60, 120, 160 & 200          \\
SimpleCNN  & CIFAR10  & Adam & 1e-3 & -   & 1e-4 & -        & -            & 100          \\
RN20 no BN & CIFAR10  & Adam & 1e-3 & -   & 0    & 0.1      & 80, 120      & 160          \\
RN20 no BN & CIFAR100 & Adam & 1e-3 & -   & 0    & 0.1      & 80, 120      & 160          \\
WRN-16-4   & CIFAR10  & SGD  & 1e-1 & 0.9 & 5e-4 & 0.2      & 60, 120      & 160          \\ 
WRN-16-4   & CIFAR100  & SGD  & 1e-1 & 0.9 & 5e-4 & 0.2      & 60, 120      & 160          \\ 
\hline
\end{tabular}
\caption{Training details for different experiments. For the Adam optimizers, the default hyperparameters were used. There was no learning rate decay for SimpleCNN.}
\label{tab:training-deets}
\end{table}

\subsection{Extended training time and learning rate ablations}

In \cref{fig:taylorized_network_training_accuracies,fig:nonlinear_advantage}, we train networks with a lower learning rate to compare to linearized training and measure the nonlinear advantage. This is done as follows: we first train a parent network at a constant learning rate of 0.1. At certain epochs (the x-axis of \cref{fig:taylorized_network_training_accuracies} and labels of \cref{fig:nonlinear_advantage}) we spawn a child network and train it with a learning rate of 0.001 and independent SGD noise until convergence (the learning rate is chosen as the smallest learning rate at which the network converges in a reasonable amount of time, 1000 epochs in our case.) We then use the learning rate dropped accuracy is the final accuracy that the child converges to. For stability, we repeat this process and average the two independent runs. The linearized training is performed using the \citet{neuraltangents2020} package and implemented in JAX \citep{jax2018github}. It trains for additional 200 epochs at learning rate 0.001 that we choose based on a small-scale grid search for linearized training specifically. The nonlinear advantage is computed as the error difference between the final test performance of the low learning rate child network, and the final test performance of the linearized neural network.

\subsection{Linearized training details}
The linearized training was performed using the Taylor expansions tools described in \citet{novak2019neural} and implemented in \citet{neuraltangents2020}. We trained for 200 epochs using SGD with Momentum, implementing the training loop in JAX \citep{jax2018github}. The learning used was $0.001$. We did a small-scale grid search for the best performing learning rate between $0.0001$ and $0.1$, choosing $0.001$ in the end.

\end{document}